\documentclass[journal]{IEEEtran}

\usepackage[utf8x]{inputenc}
\usepackage{graphicx}
\usepackage{caption}
\usepackage{subcaption}
\usepackage{amsmath,amssymb,amsfonts}
\usepackage[hyperfootnotes=false]{hyperref}
\usepackage{array}
\usepackage{float}
\usepackage{adjustbox}

\newcommand{\B}{\fontseries{b}\selectfont\small}

\hypersetup{
    colorlinks,
    linkcolor=[rgb]{0.1,  0.4,  0.7},
    citecolor=[rgb]{0.0,  0.75, 0.0},
    urlcolor= [rgb]{0.6,  0.0,  0.6}
}

\begin{document}


\title{Real-Time Human Pose Estimation on a Smart Walker using Convolutional Neural Networks}







\author{
    \IEEEauthorblockN{
        Manuel~Palermo*\thanks{* Corresponding Author.}\IEEEauthorrefmark{2},
        Sara~Moccia\IEEEauthorrefmark{3}\IEEEauthorrefmark{4},
        Lucia~Migliorelli\IEEEauthorrefmark{5},
        Emanuele~Frontoni\IEEEauthorrefmark{5},
        Cristina~P.~Santos\IEEEauthorrefmark{2}
    }\\
    \thanks{\IEEEauthorrefmark{2}{Center for Microelectromechanical Systems, University of Minho, Portugal}}
    \thanks{\IEEEauthorrefmark{3}{The BioRobotics Institute, Scuola Superiore Sant'Anna, Pisa, Italy}}
    \thanks{\IEEEauthorrefmark{4}{Department of Excellence in Robotics and AI, Scuola Superiore Sant'Anna, Pisa, Italy}}
    \thanks{\IEEEauthorrefmark{5}{Department of Information Engineering, Università Politecnica delle Marche, Ancona, Italy}}
    \thanks{E-mail addresses: a76886@alunos.uminho.pt (Manuel Palermo), sara.moccia@santannapisa.it (Sara Moccia), l.migliorelli@pm.univpm.it (Lucia Migliorelli), e.frontoni@univpm.it (Emanuele Frontoni), Cristina P. Santos (cristina@dei.uminho.pt)}
}

\maketitle

\begin{abstract}

Rehabilitation is important to improve quality of life for ~mobility-impaired patients. Smart walkers are a commonly used solution that should embed automatic and objective tools for data-driven human-in-the-loop control and monitoring. However, present solutions focus on extracting few specific metrics from dedicated sensors with no unified full-body approach. We investigate a general, real-time, full-body pose estimation framework based on two RGB+D camera streams with non-overlapping views mounted on a smart walker equipment used in rehabilitation.
Human keypoint estimation is performed using a two-stage neural network framework. The 2D-Stage implements a detection module that locates body keypoints in the 2D image frames. The 3D-Stage implements a regression module that lifts and relates the detected keypoints in both cameras to the 3D space relative to the walker. Model predictions are low-pass filtered to improve temporal consistency. A custom acquisition method was used to obtain a dataset, with 14 healthy subjects, used for training and evaluating the proposed framework offline, which was then deployed on the real walker equipment.
An overall keypoint detection error of 3.73 pixels for the 2D-Stage and 44.05mm for the 3D-Stage were reported, with an inference time of 26.6ms when deployed on the constrained hardware of the walker.
We present a novel approach to patient monitoring and data-driven human-in-the-loop control in the context of smart walkers. It is able to extract a complete and compact body representation in real-time and from inexpensive sensors, serving as a common base for downstream metrics extraction solutions, and Human-Robot interaction applications. Despite promising results, more data should be collected on users with impairments, to assess its performance as a rehabilitation tool in real-world scenarios.

\end{abstract}




\IEEEpeerreviewmaketitle



\section{Introduction}
\label{sec:intro}

Gait and Posture disabilities are common \cite{WHO2011, Mikolajczyk2018}, and increasing due to the aging population and to the global incidence of cardiovascular and/or neurological disorders, such as cerebellar ataxia, cerebral palsy and Parkinson’s disease, among others \cite{Jonsdottir2017, Johnson2016}. Along with cognitive impairments, individuals with disability may present a lack of stability, affected motor coordination, poor balance, and muscle weakness, leading to an increased risk of falls and fall-related morbidity \cite{Mikolajczyk2018, Moreira2019}.

Rehabilitation is traditionally conducted by physicians and therapists over long periods of time, demanding high physical effort from rehabilitation professionals with challenges due to variability in clinical evaluation, while being time-consuming and prone to errors due to clinicians’ fatigue \cite{Mikolajczyk2018}. Robotics-based rehabilitation is an evolving area that aims to improve the quality of life of motor-impaired people by providing residual motor skills recovery based on repetitive and intensity-adapted training. Along with assistive devices, smart walkers became a popular choice in the context of gait rehabilitation \cite{Martins2012smartwalkers, Mikolajczyk2018}.

\begin{figure}[t]
    \centering
    \includegraphics[width=0.5\textwidth]{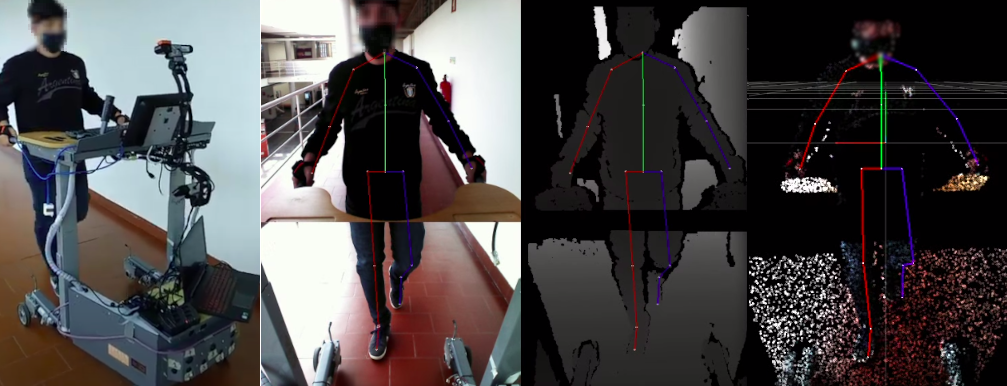}
    \caption{From left to right: Outside view of the acquisition setup; collected data from concatenated RGB frames overlaid with projected 2D skeleton; Concatenated depth frames overlaid with projected 2D skeleton; merged pointcloud overlaid with 3D skeleton (data from the gait camera is transformed, through an extrinsic transformation, to the posture camera's referential).}
    \label{fig:database_samples}
\end{figure}

An automatic and complete spatio-temporal representation of the patients' configuration in space, based on data gathered from built-in sensors, would be highly desirable to provide personalized care \cite{Kidzinski2020analysis, Chen2020Pose}. It would allow the extraction of quantitative parameters to monitor and help rehabilitation professionals evaluate patient improvements, simultaneously serving as a basis for downstream human-robot interactions and user-centered control strategies, adjusting to the patient needs in real-time.

Available smart walker solutions for patient monitoring have focused on extracting narrow aspects, such as specific gait parameters, using specialized hardware and traditional software, with no full-body detection, and presenting fundamental limitations when dealing with non-ideal conditions \cite{neto2015walkers}. Moreover, body detection errors are reported qualitatively, with no general validation scheme, limiting comparison across works \cite{neto2015walkers, Paulo2017AIWALKER, Moreira2019}.

This work addresses these challenges and contributes with a novel solution, based on current advances in human pose estimation (HPE) using Deep Learning (DL) approaches. Specifically, we propose a non-invasive framework for an accurate, real-time, and lightweight full-body HPE deployed on the ASBGo smart walker \cite{Moreira2019}, using visual information coming from both cameras mounted on the equipment. 

This work explores all steps, starting from data acquisition, model training, benchmarking, and, finally its deployment on the constrained walker hardware.

\subsection{Solution Requirements}
\label{sec:requirements}

To be effectively used as a rehabilitation tool on the ASBGo walker, a HPE solution has to:

\begin{itemize}
    \item Be able to extract a low-dimensional representation of the human body, specifying the location of the following keypoints in space: wrists, elbows, shoulders, neck, middle of the spine, pelvis, hips, knees, heels, and toes of the feet. These are considered necessary for downstream tasks, without the use of visual markers or wearables on the patient, as this is not practical when dealing with movement impaired subjects, thus resorting solely on visual information coming from the cameras.
    
    \item Run in an online fashion, since the output representation should be available promptly for downstream applications in personalized assistance and health monitoring. A sampling frequency above 19Hz ($<$53ms) is also required, as to not produce aliasing of any movement components during walking \cite{Angeloni1994gaitfreq}.

    \item Present the lowest possible detection error, where a maximum error threshold of 75mm for each keypoint was targeted. This value is half the threshold commonly used in the HPE literature  \cite{Chen2020Pose} and was discussed with a physician.
\end{itemize}

\subsection{Related Work}
Approaches to patient monitoring using smart walker devices have focused on extracting specific gait parameters \cite{neto2015walkers, Moreira2019} from dedicated sensors. To our knowledge, currently, no smart walkers present global solutions for full-body gait/posture evaluation. Moreover, detection errors are reported qualitatively, making them difficult to compare across studies.

\cite{Frizera2011Simbiosis} used ultrasound sensors placed facing each of the patients' legs, to obtain a signal used to measure gait cadence. \cite{Sierra2019AGoRA} also predicted this metric, but from force sensors on the handlebars by measuring the force on each handlebar caused by the body sideways displacement.

\cite{mou2012CAIROW} have resorted to using laser rangefinders aimed at both legs, to obtain the shank locations on a 2D plane parallel to the ground.

Pointcloud data, obtained from a depth sensor pointed at the subjects' legs, was used by \cite{Paulo2017AIWALKER} along with traditional Computer Vision (CV) techniques (clustering, Hough transform) to detect the feet locations and knee, hip and ankle kinematics.

\cite{Moreira2019}, in the current version of the ASBGo walker, have also resorted to traditional CV techniques based on depth and RGB frames obtained from two independent cameras and systems. One used to monitor the subjects feet and legs and another pointed at the chest for the posture, yielding multiple full body metrics.

Although useful, all the above methods suffer from fundamental flaws when dealing with: non-ideal conditions (\textit{e.g.},  feet occlusions), variable positional offsets depending on body segments' thickness or use of model assumptions, relying on multiple sub-systems for the different body parts with no unified full-body approach, and thus not exploiting existing movement dependencies.

DL algorithms have shown great potential for HPE \cite{Chen2020Pose, mehta2017vnect}, being capable of providing custom-fit solutions for the problem requirements, with low detection error, robustness to environmental conditions while using cheap consumer hardware \cite{Groos2020efficient, mehta2017vnect}. Nevertheless, these require large amounts of labeled data to achieve good performance, which, in the case of 3D keypoint positions, is often not trivial and time-consuming to obtain \cite{Chen2020Pose}, and most methods were not developed for real-time applications \cite{Cao2017openpose, mehta2017vnect, Groos2020efficient}.

An automated solution for full body analysis in rehabilitation scenarios was proposed by \cite{Kidzinski2020analysis}. OpenPose \cite{Cao2017openpose} is used to infer the patients' keypoints on video over a rehabilitation session. Features are then extracted and fed to a regression model to produce gait metrics, with good correlation to physicians' reports. However, the solution does not run in real-time.

\cite{Moccia2020preterm}, used a spatio-temporal CNN to detect limbs of preterm infant babies in neonatal intensive care units. A camera was placed above the infant's crib, automatically detecting its movement at all times to monitor health conditions, without requiring medical supervision. A similar idea was explored by \cite{Achilles2016blanket}, by placing cameras over patients in hospital beds, to automatically detect their pose and alert to complications.

In this work, we present a non-invasive framework for accurate and real-time full-body patient pose estimation, leveraging current breakthroughs in the HPE literature using DL, which will be deployed on the ASBGo walker used in patient rehabilitation.

Section \ref{sec:methods} describes the equipment used in this study, as well as the methods employed from the dataset acquisition, to NN model training and deployment on the walker. The experimental details, considerations, and protocols necessary, to replicate this study, are specified in Section \ref{sec:experimental}.
Section \ref{sec:results} presents all the results obtained with the proposed method, as well as ablation studies of some components, which are then discussed in Section \ref{sec:discussion}, along with the limitations, while offering multiple insights with connections to the literature.
Section \ref{sec:conclusion} concludes the article, with a short synthesis of the work developed, along with directions for future research.

\section{Methods}
\label{sec:methods}

\begin{figure*}[t]
    \centering
    \includegraphics[width=\textwidth]{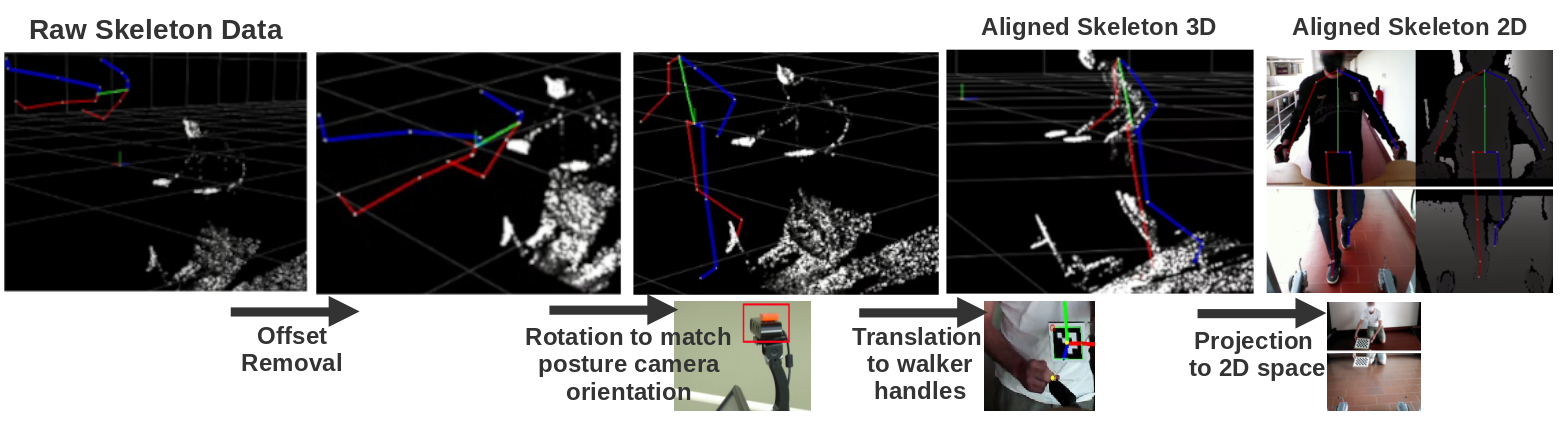}
    \caption{Transformations to spatially align the Xsens skeleton data from a world referential (left) to the walker's posture camera referential (right). Transformations are shown in 3D space along with the 3D pointcloud for reference. Once the skeleton is aligned in 3D space, it is possible to project it to the cameras' referential.}
    \label{fig:methods_dataset_SkeletonAlignProcessingSteps}
\end{figure*}

This section describes the equipments used in this work, as well as the methods to acquire the custom dataset and to pre-process the data. This is followed by the description of the model framework and the developed NN architecture. This section ends with the description of the post-processing steps and deployment strategy on the smart walker.

\subsection{Materials}
\subsubsection{ASBGo smart walker} The ASBGo smart walker \cite{Moreira2019}, used in rehabilitation, will run the proposed HPE framework. It is equipped with an Intel NUC-6i7KYK (Intel Corporation, The United States) mini-computer (Intel i7 4-core 2.60GHz CPU, 8GB RAM), responsible for all the high-level algorithms and GUI, while communicating with multiple sensors used for status, patient and environment monitoring, using a ROS 1 \cite{Quigley2009Ros} messaging interface. During rehabilitation sessions it is tasked with running multiple processes concurrently, and has to keep responsive at all times, constraining the use of processing-heavy applications. This is challenging considering the lack of any hardware accelerator (\textit{e.g., GPU}).

Two Orbbec Astra (Orbbec 3D Technology International Inc. The United States) RGB+D cameras are used to monitor the user and are mounted on the front of the walker, in a configuration with complementing non-overlapping views. The upper camera (posture) only visualizes the upper part of the body, while the bottom camera (gait) only visualizes the legs and feet. Each obtains RGB and also depth images with a resolution of 640x480 pixels at 30fps. The depth sensor has a range of [0,10]m (errors increase with the distance from the sensor).

\subsubsection{Xsens MTw Awinda} The Xsens MTw Awinda (Xsens Technologies B.V., The Netherlands) inertial MoCap system was used to acquire GT data from the subjects using the walker. It is composed of 17 wearable IMU sensors, which communicate over wireless with a base module connected to a computer. The proprietary Xsens MVN software uses the IMU data to drive a biomechanical model of the subject, from which accurate positional and kinematic data are extracted.

\subsection{Walker Dataset}

A custom dataset was acquired to train and validate the HPE algorithms: it relates RGB+D images coming from the walker cameras, with ground-truth (GT) keypoint data (referred to as skeleton) coming from the Xsens system.

A hardware trigger was used to start the acquisition on both systems and the data were later synchronized offline using timestamps saved during recording.

The skeleton data were transformed to the posture camera's referential. First, the skeleton was centered on the origin of the referential and rotated based on the orientation of an additional Xsens IMU placed on top of the posture camera. Finally, a translation was applied, which places the skeleton wrists on the corresponding walker handles relative to the camera. This translation was obtained through extrinsic calibration using visual markers. The process is summarized in Figure \ref{fig:methods_dataset_SkeletonAlignProcessingSteps}.

With the skeleton aligned in 3D space and by knowing the transformation between the cameras, along with the cameras' intrinsics, it is possible to obtain the 2D keypoint locations in each of the camera frames by projecting the skeleton to 2D using the Pinhole Camera model. This process was used to obtain keypoint GT labels for the RGB and depth frames for both cameras. The inverse process can also be used to project 2D information from each image to the 3D space.

\begin{figure}[t]
    \centering
    \begin{subfigure}[b]{0.11\textwidth}
        \centering
        \includegraphics[height=3.75cm]{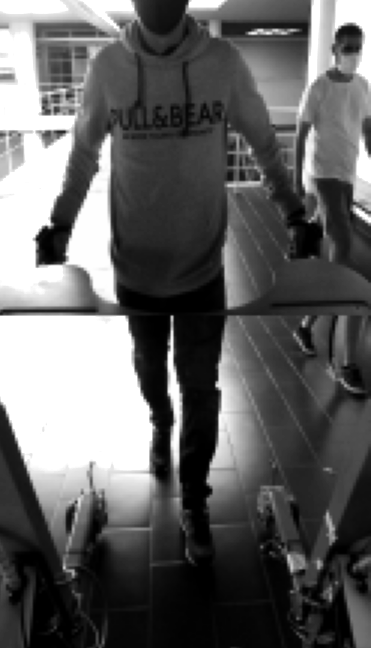}
        \caption{}
        \label{fig:methods_processing_input_image}
    \end{subfigure}
    \begin{subfigure}[b]{0.11\textwidth}
        \centering
        \includegraphics[height=3.75cm]{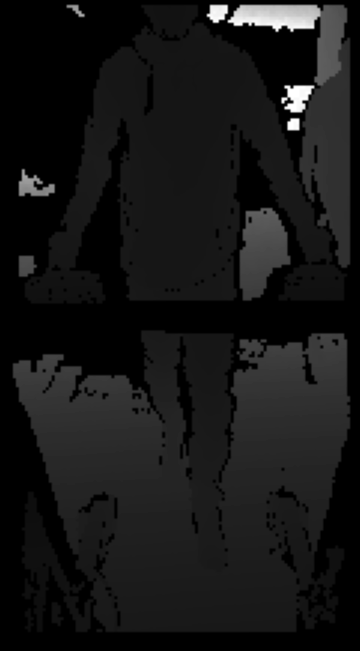}
        \caption{}
        \label{fig:methods_processing_input_depth}
    \end{subfigure}
    \begin{subfigure}[b]{0.11\textwidth}
        \centering
        \includegraphics[height=3.75cm]{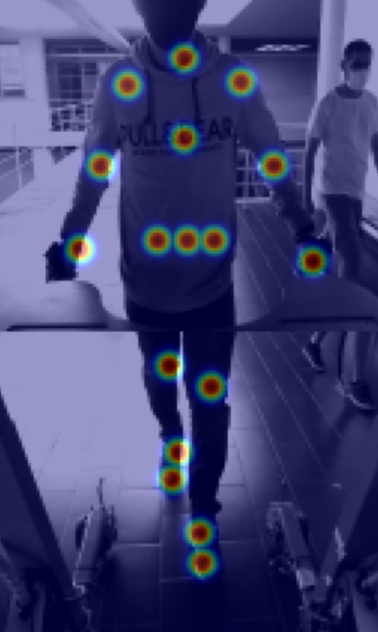}
        \caption{}
        \label{fig:methods_processing_keypointheatmaps}
    \end{subfigure}
    \begin{subfigure}[b]{0.11\textwidth}
        \centering
        \includegraphics[height=3.75cm]{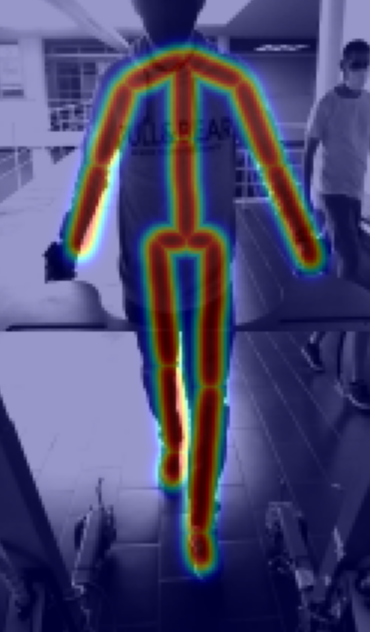}
        \caption{}
        \label{fig:methods_processing_connectionheatmaps}
    \end{subfigure}
    \caption{Processed input Image (\textbf{a}) and Depth (\textbf{b}) frames which will be fed to the model. Stacked Gaussian probability Keypoint (\textbf{c}) and Connection (\textbf{d}) heatmaps.}
    \label{fig:methods_data_preprocessing_traindata_(inputs_hmaps)}
\end{figure}

\subsection{Dataset Preparation}
\label{sec:dataset_preparation}

All data were down-sampled from the original 30Hz to 10Hz when training the models, to reduce the number of similar samples, which add little additional information to the training set.

\begin{figure*}[t]
    \centering
    \includegraphics[width=\textwidth]{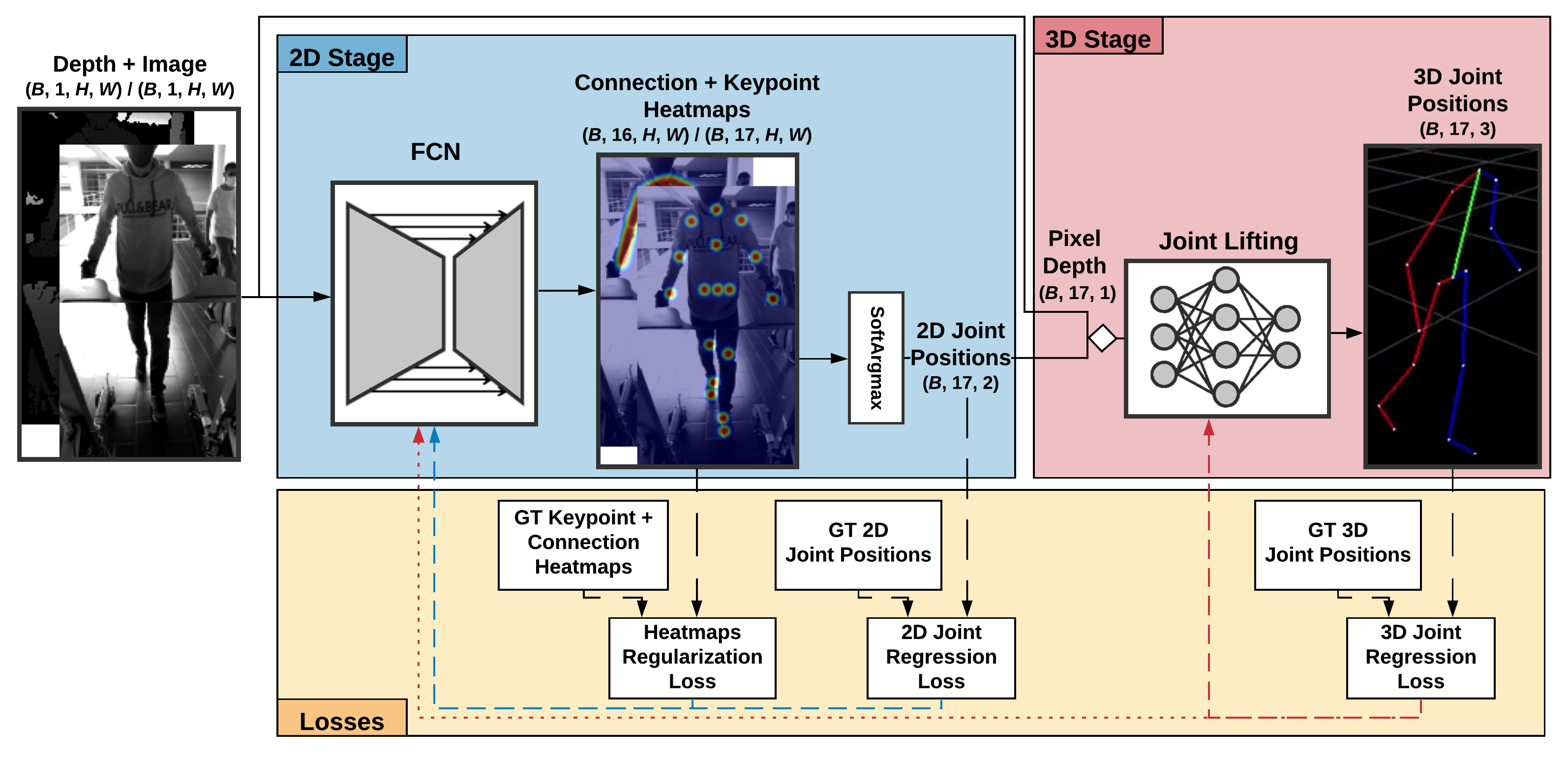}
    \caption{Proposed two-stage model framework. The 2D-Stage takes the input frames and regresses the keypoint and connection heatmaps using a FCN. Soft-argmax is used on the keypoint heatmaps to obtain the 2D keypoint locations, which are lifted to 3D space using a FC regression network aided by the depth information. A keypoint loss is applied to both 2D and 3D keypoint predictions, and the heatmaps are regularized by an intermediate loss. Shapes of the data through the network are also specified.}
    \label{fig:methods_ModelPoseEstimationDiagram}
\end{figure*}

\subsubsection{Input Frames}
\label{sec:inputs_processing}
The data used to feed HPE algorithms were pre-processed by:
\textbf{i)} RGB Frames from both cameras were converted to grayscale and normalized to [0,1] range. The depth frames, were divided by the maximum range (10m), obtaining data between [0,1]. Normalization was preferred over standardization as it preserves the meaning of the depth values.
\textbf{ii)} Posture and gait camera frames were concatenated to create a single frame with information from both parts of the body. This method was chosen over processing frames from each camera individually or doing feature fusion inside the model, as this way positional relationships can be exploited, while decreasing computational costs of processing the two frames independently. This was only possible due to the complementing views obtained from the camera placement on the walker. Some samples of the pre-processed data can be seen in Figs. \ref{fig:methods_processing_input_image}, \ref{fig:methods_processing_input_depth}).
\textbf{iii)} The frames' resolution was reduced, decreasing computation and memory requirements and thus inference time. Since the subject will always be close to the camera while using the walker, the loss of fine details should not cause a decrease in performance, while on the other hand, increasing the percentage of the frame present in the effective receptive field (ERF) of the model, without requiring a very deep architecture. Finally, 2 input features, one for the depth and another for the grayscale image, with shapes $(1, H, W)$ were obtained.

\subsubsection{Keypoint Selection}
\label{sec:inputs_joint_selection}
A subset of 17 keypoints was selected from the original Xsens skeleton obtained in the data acquisition step, following the requirements (Sec \ref{sec:requirements}). A set of 16 connections were also defined between these keypoints, following natural limb segments of the skeleton. The keypoints and connections can be seen in Figure \ref{fig:database_samples}.

\subsubsection{2D Heatmaps}
\label{sec:heatmaps_creation}
Gaussian probability heatmaps were created from the 2D keypoint locations. These will be used as intermediate regularization for the model, following the HPE literature \cite{Chen2020Pose, Newell2016hourglass, tompson2015heatmaps}. Keypoint heatmaps were created for each of the keypoints, by a Gaussian probability function centered at each 2D location and with a variance ($\sigma$) of 3 \cite{tompson2015heatmaps}. Similarly, connection heatmaps were created between connected keypoints by using 1D Gaussian distributed values along a connection line between two keypoints \cite{Moccia2020preterm}. The stacked heatmap output features for each type can be visualized in Figure \ref{fig:methods_data_preprocessing_traindata_(inputs_hmaps)}b). This creates data samples with shapes $(17, H, W)$ and $(16, H, W)$ respectively for the keypoint heatmaps and connection heatmaps.


\subsection{Model Framework}
\label{sec:model_framework}

\begin{figure*}[t]
    \centering
    \includegraphics[width=\textwidth]{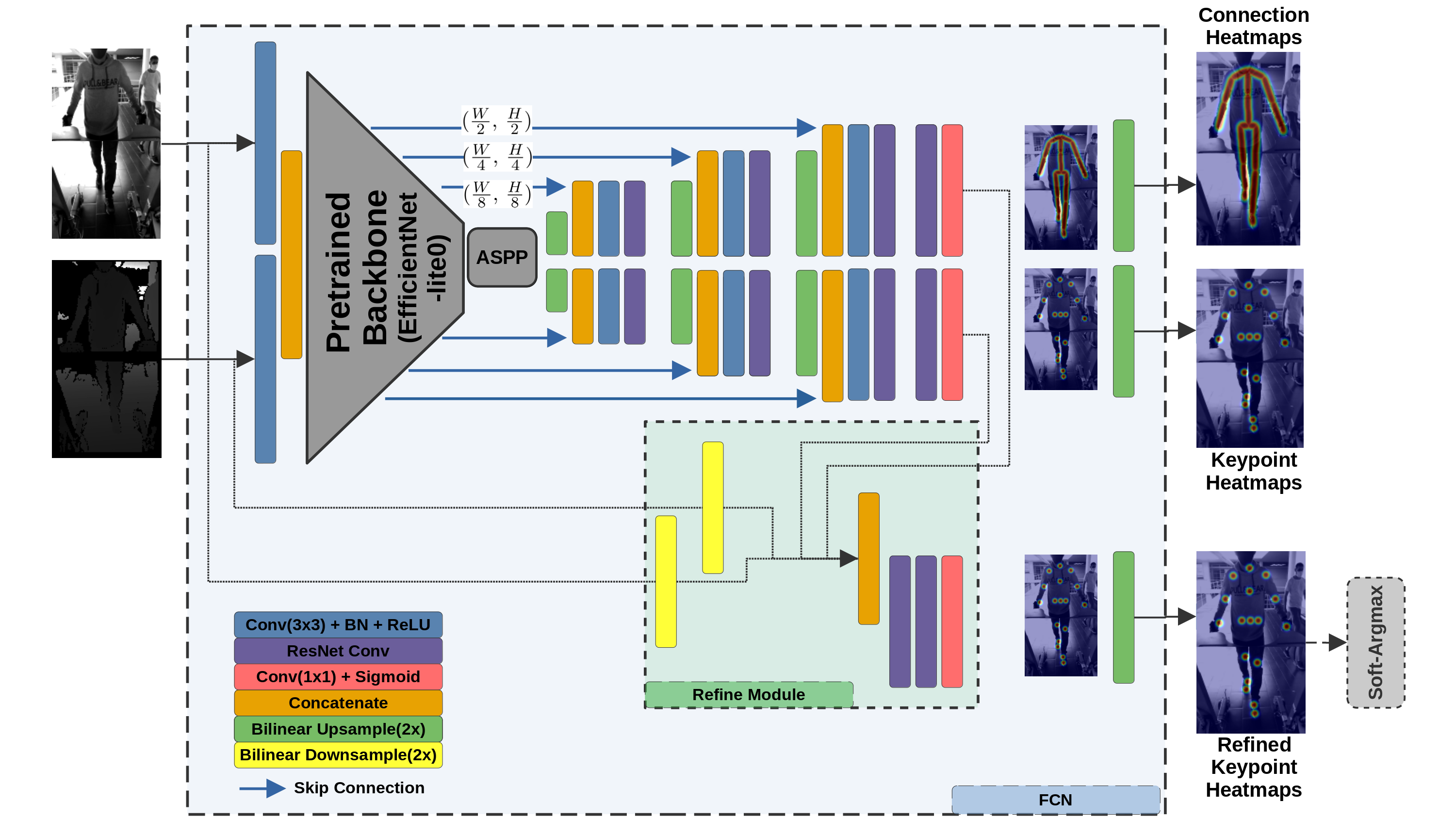}
    \caption{2D-Stage model architecture: The processed input image and depth frames are concatenated and passed by a pretrained backbone (EfficientNet-lite0), yielding multi-level features (($\frac{W}{2}$, $\frac{H}{2}$) to ($\frac{W}{16}$, $\frac{H}{16}$)), where W and H are the image width and height, respectively. The higher level features ($\frac{W}{16}$, $\frac{H}{16}$) are enhanced by an ASPP module and then up-sampled through two branches, to produce connection and keypoint heatmaps with ($\frac{W}{2}$, $\frac{H}{2}$) resolution. A refining module is used to improve keypoint heatmaps, from which the joint locations are extracted using the soft-argmax operator.}
    \label{fig:methods_model_fcn_Architecture}
\end{figure*}

The DL model framework for HPE is shown in Figure \ref{fig:methods_ModelPoseEstimationDiagram} and follows a two-stage approach. The 2D-Stage is responsible for detecting keypoints and keypoint-connections in 2D image space. The 3D-Stage regresses the keypoint locations to 3D space. This two-stage architecture was chosen above methods which directly compute the positions in 3D space \cite{Nibali2019margipose, mehta2017vnect}, as it is lighter to compute and has shown competitive results in the literature \cite{Martinez2017baseline}, while being easier to optimize since there is no internal conversion between image and cartesian spaces \cite{Nibali2019margipose}. Moreover, it allows dealing with the multi-camera fusion problem in separate stages without having to learn a global internal representation.

\subsubsection{2D-Stage}
\label{sec:architecture_2d}

The 2D-Stage of the model is responsible for detecting the keypoint pixel locations from the input image and depth frames. This is accomplished with a Fully Convolutional Network (FCN) based on multiple works in the literature \cite{Cao2017openpose, Newell2016hourglass, Wei2016cpm, Moccia2020preterm, Groos2020efficient, Sun2018integral, artacho2020unipose}. Special attention was given to the computational cost given the hardware constraints. The architecture is displayed in Figure \ref{fig:methods_model_fcn_Architecture}.

A backbone based on the lite variants of the EfficientNet \cite{Tan2019efficientnet} architecture was used as a general feature extractor similarly to \cite{Groos2020efficient}, settling for the lite0 version. The first 4 resolution feature blocks were selected, yielding multi-level features, with dimensions from ($\frac{W}{2}$, $\frac{H}{2}$) to ($\frac{W}{16}$, $\frac{H}{16}$).

Two decoder branches, with skip connections from the backbone intermediate features \cite{Newell2016hourglass}, up-sample these representations to produce the output keypoint and connection heatmaps respectively on each branch, with ($\frac{W}{2}$, $\frac{H}{2}$) resolution.
The bottleneck features ($\frac{W}{16}$, $\frac{H}{16}$) were enhanced using an Atrous Spatial Pyramid Pooling module (ASPP) \cite{Chen2018deeplab}, to increase the ERF of the modelsimilarly to \cite{artacho2020unipose}.

The bias of the last convolution layers before each output heatmap is initialized with the method proposed by \cite{Lin2017focal}. This decreases the probability of keypoint detection in each pixel, reducing early training instability due to the inherent class imbalance associated with the sparse heatmap values.

A shallow refining module is used to improve the keypoint heatmaps, by aggregating information from both heatmap branches, along with 2x down-sampled input frames \cite{Moccia2020preterm}. The computation is performed in half resolution size to decrease the computational cost, and the output heatmaps are up-sampled to the input frame size through bilinear interpolation, following \cite{Newell2016hourglass, artacho2020unipose}.

Finally, each keypoint location in the input frames is extracted from the refined keypoint heatmaps using the \textit{soft-argmax} operator \cite{Sun2018integral}, along with the detection confidence.

\subsubsection{3D-Stage}
\label{sec:architecture_3d}

The 3D-Stage consists of the residual linear model proposed by \cite{Martinez2017baseline}, lifting the 2D keypoint locations from the previous stage to 3D. The 2D keypoint locations are previously normalized, converting them from pixel locations to normalized pixel space, while maintaining the aspect ratio. This increases training stability and generability to different input frame resolutions. Additionally, depth information at each pixel location was given to help resolve pose ambiguities and ease the learned problem of point projection.

A total of 256 eurons per hidden layer were used, instead of the original 1024 \cite{Martinez2017baseline}. In our preliminary analysis, this was found to reduce overfitting while making the model faster to train. This may be explained given the smaller number of parameters (roughly 0.29M for 256 neurons compared to 4M for 1024 neurons compared to 0.29M for 256 neurons).

\subsection{Losses}
\label{sec:losses}

The 2D-Stage of the model was trained using the integral loss proposed by \cite{Sun2018integral}, where the MAE between predicted 2D keypoint positions and the corresponding GT is minimized. This is combined with a heatmap MSE regularization term applied to all heatmaps.

The 3D-Stage was trained with a Log-Cosh loss between 3D keypoint positions and corresponding GT. This loss combines the outlier robustness of the MAE with the diminished update size for smaller error of the MSE loss. This choice is supported by Gonzalez et al. which uses a similar Smooth L1 loss with superior results over the standard MSE/MAE losses \cite{Gonzalez2020residual}.

\subsection{Data Post-Processing}
\label{sec:post_processing_temporal_filter}

The predictions produced by the complete model, although coherent (if trained properly) at each frame, contain no temporal information, resulting in high-frequency positional jittering when applied over a sequence of frames, producing incorrect joint kinematics which might be incompatible with downstream applications. To tackle this issue, a similar approach to \cite{mehta2017vnect} was taken, by running a first-order low-pass adaptive filter \cite{casiez2012oneeurofilter} as a post-processing step on top of the model predictions. This filter was especially designed for high responsiveness and low computational cost, for real-time applications, showing good empirical results, even when compared to model-based alternatives (\textit{e.g.}, Kalman Filters).
Its main feature is the use of a cutoff frequency that rapidly adapts to the input signal, decreasing for slow signals to reduce jitter, and increasing for faster signals to reduce delay.

\subsection{Model Deployment}
\label{sec:model_deployment}

After offline training, the model was exported to \textit{ONNX\footnote{\url{https://github.com/onnx/onnx}}}, an open-source and widely supported framework focused on model inferencing for production. This removes most dependencies while offering optimization tools to improve runtime latency. The model was loaded to the walker's existing C++ ROS 1 (Melodic Morenia) \footnote{\url{http://wiki.ros.org/melodic}} environment using the ONNX Runtime\footnote{\url{https://github.com/microsoft/onnxruntime}} accelerator library built with the default CPU provider. All available optimizations were used which include: pre-computation of constant values, removal of redundant nodes used during training (\textit{e.g.}, dropout, identity, etc...), common node fusion into single operations, data layout, and memory access pattern optimization. All the data pre-processing steps (Section \ref{sec:inputs_processing}) were performed for each group of 4 synchronized frames (image and depth for each camera) before feeding the data to the model.

\section{Experimental Protocol}
\label{sec:experimental}

This section goes into detail about the experimental protocols, including details on the dataset creation, data pre/post-processing and networks training and evaluation. Competing network architectures are also considered at the end of the section.

\subsection{Dataset Details}
Trial conditions followed standard gait rehabilitation procedures with the walker, and were defined in collaboration with an expert physician. Each subject was instructed to perform 27 trials composed of:  3 sequences (walking forward, cornering left, cornering right); at 3 different speeds (0.3, 0.5, 0.7 m/s) \cite{Beaman2010speed}; each repeated 3 times (with the same course) in different corridors (to maximize environment variability. The final dataset contains a total of 378 trials, from 14 healthy subjects.

This amounts to 166k frames of synchronized data sampled at 30Hz (92 minutes of total recording time). Data from each subject were divided into train (9 subjects), validation (2 subjects), and test (3 subjects) splits, resulting in $\approx$110k, $\approx$19k, $\approx$36k samples for each split, respectively. A sample can be seen in Figure \ref{fig:database_samples}.

\subsection{Implementation Details}

The image and depth input frames were resized \footnote{OpenCV's resize function with Area and nearest-neighbor methods respectively: \url{https://docs.opencv.org/3.4/da/d54/group__imgproc__transform.html}}. from a resolution of 640x960 pixels\footnote{Concatenated frames have a height of $2\times480=960$ pixels.} to 128x224 ($W=128$, $H=224$), decreasing computation and memory requirements and thus inference time.

Models were trained in two steps. First, only the 2D-Stage was trained, using the image and depth frames as inputs to produce the intermediate 2D features: keypoint locations and the keypoint and connection confidence heatmaps. The 3D-Stage of the model was then trained by using as input the 2D features, with that stage's weights frozen (to prevent destroying previously learned features), and outputting the 3D keypoint positions.
End-to-end learning of the complete model was tried in early experiments, however, it led to worse 2D heatmaps and was thus dropped, also decreasing training complexity.
%

Reasonable hyper-parameters for training were found empirically and kept constant for all models tried. The Adam optimizer was selected with an initial learning rate of 2e-3 which decayed to 1e-5 over 30 epochs using a cosine-annealing schedule. A batch size of 16 and 32 was, respectively, used to train the 2D and 3D-Stages of the model. Gradient clipping with a range of [-0.2, 0.2] was applied during training for all models to prevent high gradient updates, especially on the first batches of training which could destroy some of the pre-trained weights. All convolutional and FC layers, except for the outputs, were followed by batch normalization (BN) and a Rectified Linear Unit (ReLU) non-linearity.

Random train-time data augmentation \cite{Shorten2019augm} was used to increase visual variability of the training set, decreasing over-fitting to the limited number of train samples. However, the depth frames could not be augmented with common operations (as it would produce incorrect depth information), so only pixel dropout was applied. Occlusion specific augmentation was also applied following the method by \cite{Sarandi2018augm} of adding random occlusion objects to the image frames, with dropped depth pixels. Additionally, when training only the 2D-Stage, affine transformations were applied to decrease over-fitting to frequent keypoint locations on the walker dataset. This produced slight incoherence in the depth information, but led to better overall results.

Additional train-time regularization was applied to most layers of the model in the form of dropout, with a percentage of 50\% for all the linear layers and spatial-dropout with a probability of 20\% for the convolutional layers. A L2 weight decay parameter with a value of 1e-5 was also added.

The filter's parameters were based on the ones used by \cite{mehta2017vnect} and tuned empirically to our model. The minimum cutoff frequency was set to $fc_{min}=1.5$ and the adaptive cutoff rate to $\beta=0.15$.

\subsection{Performance Metrics}

The models were evaluated considering commonly used metrics in the HPE literature \cite{Chen2020Pose}, as follows.

\textit{Mean Per-Joint Position Error (MPJPE)} - Average Euclidean distance between a GT position and a predicted position for each of the keypoints is calculated after performing root joint alignment (in this case the pelvis). It will be the most focused in this work, since most downstream patient analysis applications use root-relative joint positions.

\textit{Procrustes-Aligned MPJPE (PA\_MPJPE)} - MPJPE but Procrustes analysis is first performed, ignoring affine errors.

\textit{Absolute MPJPE (A\_MPJPE)} - Similar to MPJPE but uses absolute positional values relative to the camera.

\begin{figure}[t]
    \centering
    \includegraphics[height=3.8cm]{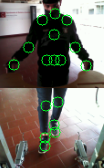}
    \caption{Chosen PCK threshold radius of 6 pixels. Detection values inside the circles, for each keypoint, are considered correctly detected.}
    \label{fig:evaluation_pck2d_threshold_6pixels}
\end{figure}

\textit{Percentage of Correct Keypoints (PCK)} - Percentage of predicted keypoints with an error below a certain threshold. 75mm was selected for all 3D tests, following the requirements in Section \ref{sec:requirements}. This metric was also evaluated in 2D where a threshold of 6 pixels was selected and can be seen in Figure \ref{fig:evaluation_pck2d_threshold_6pixels}. 

\textit{Inference Time (latency)} - Latency was evaluated on a Nvidia Tesla T4 GPU and on an Intel i7 - 4720HQ CPU. This is critical given the real-time nature of the application, and refers to the time necessary to load the inputs into the device and do a forward pass for a single sample. The results were obtained by running inference on all samples in the test split.

\subsection{Model Variants}
\label{sec:model_variants}

Alternative versions of the 3D-Stage were also considered, based on ideas presented in the literature, and in a search for the best configuration:

\subsubsection{Baseline} Uses the original model proposed by \cite{Martinez2017baseline}. It is similar to the default 3D-Stage method but without including the depth information as input to perform the lifting and considering the original 1024 channels per layer.

\subsubsection{Semantic Graph Convolutions (SemGCN)} Method proposed by \cite{Zhao2019semgcn} was tried, with the addition of depth information for each keypoint as input. This module exploits the hierarchical structure of the skeleton by using state-of-the-art graph convolutions that aggregate information along connected joints. The non-local layers used in the original work were not used here as they doubled the inference time without noticeable improvements.

\subsubsection{Projection Residual} Follows the approach by \cite{Gonzalez2020residual}, extending it to a non-overlapping multi-camera setup. Instead of learning to project the data from 2D space to 3D similarly to the lifting methods, an explicit projection is computed and then refined, as follows.

The detected 2D keypoints in each of the camera frames are first projected to 3D space relative to the posture's camera referential, using the depth information and the intrinsic parameters for each of the cameras, based on a Pinhole camera model. Then it is applied an extrinsic transformation to the projected gait camera data, so it is in the posture camera's referential, this produces a rough estimate of the detected keypoint positions, but affected by fundamental flaws: \textbf{i)} missing depth information due to dead pixels in the depth frame, which occurs quite frequently, and make it impossible to project the affected keypoints to 3D space; \textbf{ii)} the person's body thickness, which will produce a varying offset for each keypoint and will be highly dependent on the subject using the equipment; \textbf{iii)} projection of incorrect pixels due to bad detections in 2D space, which can lead to unreasonable keypoint locations, especially when this occurs for points in the background; \textbf{iv)} cannot deal with keypoint occlusions, since these would be projected incorrectly to 3D space; \textbf{v)} small random error dependent on the depth sensor noise, which reduces temporal coherence.

The 3D flawed transformation can be improved by applying a series of residual FC layers to produce a globally coherent result by using spatial information from the other keypoints.

This method provides a more principled way to fuse the depth information from both cameras. An explicit transformation (which can be obtained through extrinsic calibration with low error) is used instead of relying on a learned internal representation that would need retraining for different camera spatial arrangements.

\subsubsection{Spatio-Temporal Features}
\label{sec:model_variant_spatiotemporal}

A sequential frame approach was also tried. It aggregates temporal information from multiple frames, following works from \cite{Moccia2020preterm, Pavllo2019temporal} and is achieved by stacking $S$ successive frames which are processed using temporal convolution blocks. These not only aggregate local spatial information, but also temporal information. $S=4$ was picked, as bigger sequential values would require significant computational requirements, especially for the 2D-Stage, while being harder to train, and to implement in real-time (since it would require buffering multiple frames before executing a prediction).

All received frames are processed simultaneously, yielding corresponding predictions for each one. In the case of the 2D-Stage, the 2D convolutions are replaced with 3D convolutions, while in the case of the 3D-Stage, the FC layers are replaced with 1D convolutions, where the extra dimension represents time.

The backbone in the 2D-Stage was also replaced with a similar temporal MobilenetV2 \cite{Sandler2018mobilenet2} architecture (since no pre-trained temporal EfficientNet version was found). The low-level layers also have temporal pooling operations removed (spatial pooling is still performed), as temporal invariance is not desirable since a sequential prediction for all 4 frames is necessary.

\section{Results}
\label{sec:results}

Results for both the 2D- and 3D-Stages of the model were evaluated on the walker dataset in the following section. Unless otherwise stated, the results only contemplate the raw model outputs, without the temporal filter post-processing.

\subsection{2D-Stage}

\begin{figure}[t]
    \centering
    \begin{subfigure}[b]{0.10\textwidth}
        \centering
        \includegraphics[height=3.8cm]{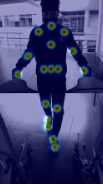}
        \caption{}
        \label{fig:results_only2d_jointheatmaps}
    \end{subfigure}
    \begin{subfigure}[b]{0.10\textwidth}
        \centering
        \includegraphics[height=3.8cm]{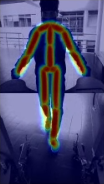}
        \caption{}
        \label{fig:results_only2d_connectionheatmaps}
    \end{subfigure}
    \begin{subfigure}[b]{0.10\textwidth}
        \centering
        \includegraphics[height=3.8cm]{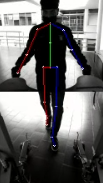}
        \caption{}
        \label{fig:results_only2d_skeleton}
    \end{subfigure}
    \begin{subfigure}[b]{0.15\textwidth}
        \centering
        \includegraphics[height=3.8cm]{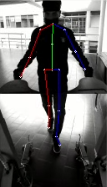}
        \caption{}
        \label{fig:results_only2d_labelskeleton}
    \end{subfigure}
    \caption{\textbf{(a)} Keypoint heatmaps, \textbf{(b)} Connection heatmaps and \textbf{(c)} 2D keypoints and connections predicted by the 2D-Stage of the model. \textbf{(d)} Corresponding 2D keypoint GT labels. All data are overlaid on top of the input image frame.}
    \label{fig:results_only2d_detection_good}
\end{figure}

The 2D-Stage (Section \ref{sec:architecture_2d}) was evaluated in the following section. Examples of features obtained for a frame of one of the test subjects can be seen in Figure \ref{fig:results_only2d_detection_good}. The model not only produces the 2D keypoint locations, but also the refined keypoint heatmaps used to obtain them, and the connection heatmaps, these are compared against the GT skeleton from the Xsens.

\begin{figure}[t]
    \centering
    \includegraphics[width=0.5\textwidth]{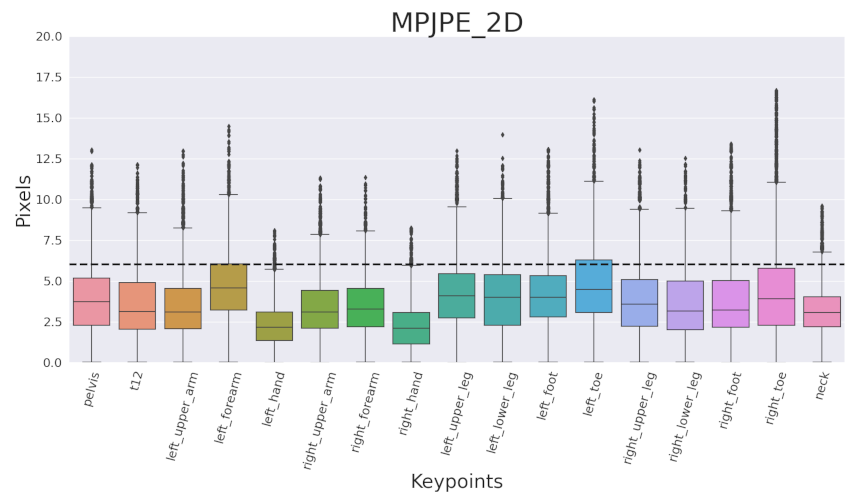}
    \caption{Boxplot per-joint error for the 2D detection, across all test frames obtained from the 2D-Stage (extreme outliers were removed for better visibility). The dashed line marks the 6 pixel threshold defined.}
    \label{fig:results_only2d_boxplot}
\end{figure}

The detection error for each keypoint was further analysed through a boxplot graph depicted in Figure \ref{fig:results_only2d_boxplot}. All keypoints display a relatively similar detection error, with a mean of 3.76 pixels, corresponding to 85.27\% of detections being below the chosen detection threshold of 6 pixels,
with an inference time of 11.97ms and 37.56ms in the GPU and CPU respectively.

\subsection{Complete Model}

\begin{figure}[t]
    \centering
    \includegraphics[width=0.5\textwidth]{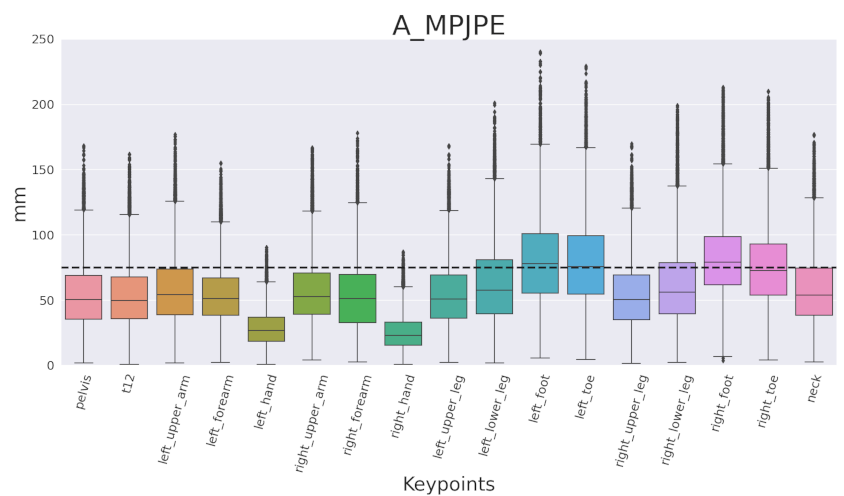}
    \caption{Boxplot per-joint error results for the 3D detection, across all test frames, with absolute values relative to the posture camera referential, obtained with the complete model (extreme outliers were removed for better visibility).}
    \label{fig:results_fullmodel_boxplots}
\end{figure}

Figure \ref{fig:results_fullmodel_boxplots} depicts the 3D error for each keypoint for the complete model (Section \ref{sec:model_framework}) relative to the camera. An average absolute error of 59.5mm relative to the posture camera was obtained, corresponding to 73.1\% of correct detections. A root-relative error of 44.1mm was obtained, with a PCK of 83.0\%. The feet keypoints displayed the largest mean errors, closer to the imposed detection threshold.
After applying a Procrustes transformation, the error is around 33.3mm, signaling the presence of affine errors, in the form of positional offset, rotation or scale, which when removed yield a PCK of 96.3\%.

The default model was further compared with other 3D-Stage alternatives explored in Section \ref{sec:model_variants} from the predictions of the default 2D-Stage. The results are summarized in Table \ref{tab:table_complete_3D_results}. A similar performance was encountered for all models with a MPJPE $\approx 46.0mm$ and PCK of $\approx 81.4 mm$.

\begin{table*}[t]
    \centering
    \caption{Results summary of the 3D-Stage and comparisons against the different variants for regression from the 2D-Stage keypoints. The best results are highlighted in bold.}
    \maxsizebox{\textwidth}{!}{
    \begin{tabular}{l | c | c | c | c | c | c}
        Method                               &   MPJPE(mm)  & PA\_MPJPE(mm) & PCK@75(\%)   & GPU latency(ms) & CPU latency(ms) & \#Params \\ \hline \hline
        2D-Stage                             &  &  &  &  &  & \\
        \hspace{0.3cm}+ Default 3D-Stage     &\B44.05 ± 0.39&\B33.35 ± 0.29 &\B83.03 ± 0.48&\B 13.43 ± 0.03  &\B38.93 ± 0.13   &  1.34M   \\
        \hspace{0.3cm}+ Baseline             & 45.80 ± 0.40 &  34.78 ± 0.31 & 81.31 ± 0.5  &   14.68 ± 0.04  & 41.28 ± 0.10    &  5.34M   \\
        \hspace{0.3cm}+ SemGCN               & 45.85 ± 0.42 &  34.80 ± 0.31 & 81.45 ± 0.51 &   19.21 ± 0.03  & 43.86 ± 0.15    &\B1.31M   \\
        \hspace{0.3cm}+ Projection\_Residual & 48.35 ± 0.44 &  36.76 ± 0.39 & 79.88 ± 0.49 &   14.50 ± 0.02  & 39.18 ± 0.12    &  1.34M   \\
        \hline
    \end{tabular}
    }
    \label{tab:table_complete_3D_results}
\end{table*}

\begin{figure}[t]
    \centering
    \includegraphics[width=0.45\textwidth]{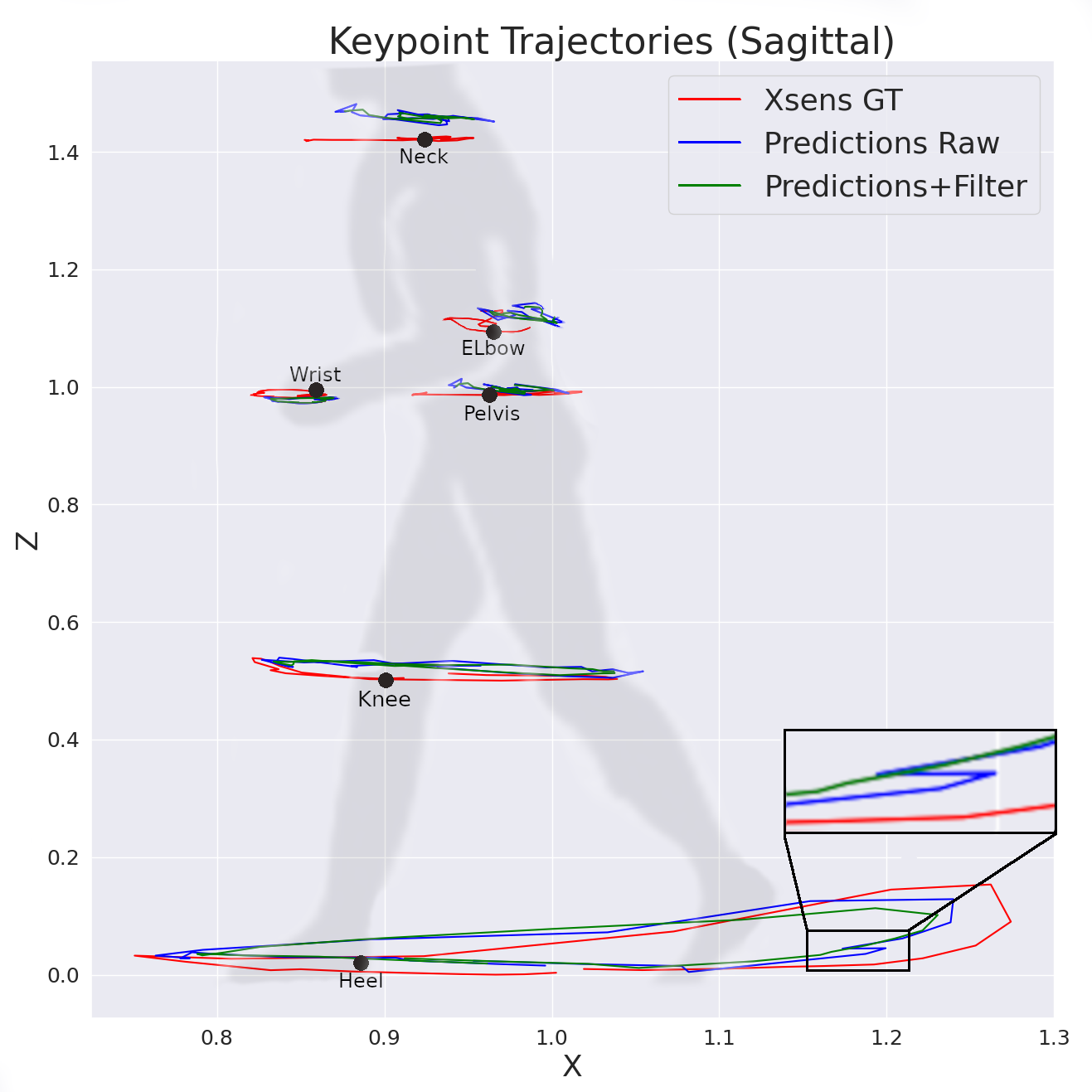}
    \caption{Trajectories described by some keypoints on the sagittal plane over a gait cycle obtained from the Xsens GT (red), Raw NN model predictions (blue) and model predictions after filtering (green). Only the neck, pelvis and left side wrist, elbow, knee and heel keypoints are shown for better visualization. The zoomed region shows the jittering trajectories produced by the raw NN predictions (blue), which can be smoothed by the filter (green).}
    \label{fig:results_3d_saggital_filter}
\end{figure}

The raw model predictions, despite showing similar overall results compared to the Xsens GT (Figure \ref{fig:results_3d_saggital_filter}), still display some constant positional offsets (\textit{e.g.}, Neck), while failing to capture the full range of motion on others (\textit{e.g.}, Heels). Moreover, the limb segment lengths produced are not constrained to match the user's, resulting in incorrect values when compared to the Xsens model (\textit{e.g.}, Wrist-Elbow).

\subsection{Temporal Filter}

The proposed model produces visually reasonable results at each frame. However, since the model has no information about previous predictions, when applied to a video sequence, it produces temporally incoherent results in the form of position jitter around the keypoint location, visible both in the 2D and 3D predictions.

The application of the low-pass filter described in Section \ref{sec:post_processing_temporal_filter} is capable of countering the high-frequency jitter. Keypoint positions obtained from the raw NN model prediction and with the temporal filter post-processing are compared in the sagittal plane over one gait cycle in Figure \ref{fig:results_3d_saggital_filter} and to the GT Xsens data.

The filter is entirely capable of running in real-time, with a processing time of 0.1ms, removing high-frequency noise from the raw model predictions, which would otherwise produce wrong joint kinematics (Figure \ref{fig:results_3d_saggital_filter} - zoomed region). However, it attenuates some natural high-frequency dynamics and introduces a slight output delay.

\subsection{Deployment}
The ONNX Runtime optimizations decreased the runtime latency of the model from the original 38.9ms to 23.1ms, with an additional 3.5ms to pre-process the input frames (Section \ref{sec:inputs_processing}), while keeping a similar detection error.

During a normal rehabilitation session, due to other concurrent systems running, the latency of the HPE framework integrated on the ROS environment was higher, with also higher variability, ranging from 25ms to 70ms (averaging 40ms), with a mean of 40ms. Moreover, results are published with some response delay (around 0.3s) given the asynchronous nature of the underlying ROS system in the equipment.

\begin{figure}[t]
    \centering
    \begin{subfigure}[b]{0.24\textwidth}
        \centering
        \includegraphics[height=5cm]{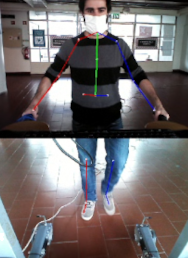}
        \caption{}
        \label{fig:results_deployed_2d}
    \end{subfigure}
    \begin{subfigure}[b]{0.24\textwidth}
        \centering
        \includegraphics[height=5cm]{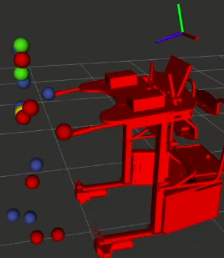}
        \caption{}
        \label{fig:results_deployed_3d}
    \end{subfigure}
    \caption{Predictions obtained from the model running on the smart walker, in 2D (\textbf{a}) and 3D (\textbf{b}) spaces. Connections between the hips and legs are not rendered in 2D due to the discontinuity between camera frames. The 3D visualization used the RViz package from the ROS environment to render the 3D keypoint locations relative to the walker. The posture camera's frame of reference is also displayed.}
    \label{fig:results_deployed_model_good_predictions}
\end{figure}

During normal walking the model performs well, being capable of detecting the 2D and 3D keypoint locations (Figure \ref{fig:results_deployed_model_good_predictions}) as expected. However, it performs sub-optimally when confronted with body configurations that lie outside the normal training distribution, for example, when walking on the sides of the walker, bringing the feet high above the ground, complete feet occlusion, to name a few.


\subsection{Benchmark and Ablation Studies}
Multiple ablation studies were conducted by removing certain components of the complete pipeline, to identify the contribution of each to the overall results. The findings are described in the next section, along with a comparison to alternative model variants (Section \ref{sec:model_variants}).

\subsubsection{Input Modalities}
\begin{figure}[t]
    \centering
    \begin{subfigure}[b]{0.475\textwidth}
        \centering
        \includegraphics[width=\textwidth]{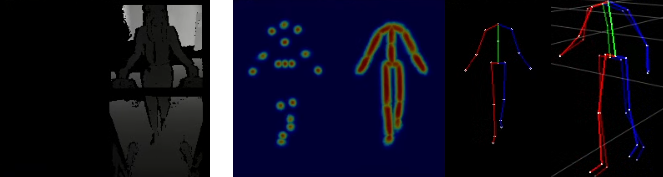}
    \end{subfigure}
    \begin{subfigure}[b]{0.475\textwidth}
        \centering
        \includegraphics[width=\textwidth]{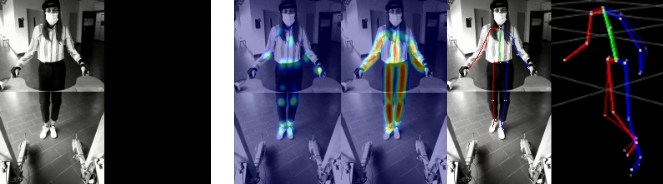}
    \end{subfigure}
    \caption{Results obtained when removing information from image (upper images) and depth (lower images) inputs. The image / depth inputs for each experiments are grouped in the left side, while the corresponding model predictions (keypoint heatmaps, connection heatmaps, 2D keypoints, 3D keypoints) are grouped in the right. 2D outputs were overlaid on the image input. 3D GT keypoints keypoints are shown overlaid with higher transparency.}
    \label{fig:results_corrupted_inputs}
\end{figure}

The proposed model is not too dependent on any of the two input modalities (image, depth). It is robust to corruption of one of the input frames, giving reasonable predictions by focusing on information from the other, even though not being explicitly trained to do so. Examples of this can be seen in Figure \ref{fig:results_corrupted_inputs}.

The model appears to be more affected by corruption of the depth input, giving low confidence detections on the heatmaps with also higher degree of keypoint error in the 2D space. Nevertheless, even without the depth information, it is still capable of regressing the positions of the 3D keypoints.

On the other hand, it seems to be less affected by corruption in the input image, as the heatmap predictions still display a well-defined shape, with high detection confidence.

Nevertheless, in both cases, a higher degree of position jittering over multiple frames is reported from the raw NN predictions.

\subsubsection{Backbone}

The chosen EfficientNet 2D backbone performance was benchmarked against commonly used ResNet architecture in Table \ref{tab:table_backbones_results}. All models achieved a similar detection error around 3.75 pixels, with slightly better results for the ResNet50 backbone (3.66 MPJPE\_2D). However, it displayed a significantly larger latency (238.9ms), being 6.36 times slower when processing in CPU compared to the default EfficientNet-lite0 option (37.56ms), which displayed the best computation time in the CPU, being faster than the light ResNet18 (94.7ms) by 2.5 times while slightly more accurate.

\begin{table*}[t]
    \centering
    \caption{The default EfficientNet-lite0 backbone 2D-Stage performance in comparison with the common ResNet models. OpenPose is also compared in terms of latency for reference (values were taken from the paper), as it is a common baseline on real-time HPE. The best results in each metric are highlighted in bold.}
    \maxsizebox{\textwidth}{!}{
    \begin{tabular}{l | c | c | c | c | c}
        Method                         & MPJPE\_2D(px) & PCK\_2D@6(\%)  & GPU latency(ms) & CPU latency(ms) & \#Params \\ \hline \hline
        Default                        &   3.73 ± 0.04 &   85.27 ± 0.59 &   11.97 ± 0.03  &\B 37.56 ± 0.13  &\B 1.05M  \\
        ResNet50                       &\B 3.66 ± 0.04 &\B 86.93 ± 0.56 &   15.27 ± 0.04  &   238.89 ± 0.66 &   26.8M  \\
        ResNet18                       &   3.88 ± 0.05 &   84.65 ± 0.61 &\B 11.88 ± 0.03  &   94.70 ± 0.35  &   12.1M  \\
        \hline
        OpenPose\cite{Cao2017openpose} &    -          &   -            & $\approx$36.00  &$\approx$10396.00&   -      \\
        \hline
    \end{tabular}
    }
    \label{tab:table_backbones_results}
\end{table*}

The commonly used OpenPose \cite{Cao2017openpose}, was also considered for the 2D-Stage, as it boasts good performance with real-time inference. However, this option was quickly dropped after checking the latency on the CPU, where the authors point to a latency around 10s for a single frame, making it unacceptable for real-time HPE on the walker hardware.

\subsubsection{Refine Module}

Only the keypoint heatmaps are necessary to extract the keypoint locations, which can be extracted directly from the heatmap branch in the case where the refine module is not used. Thus, it is possible to completely ignore the computation of the connection heatmaps branch to increase run-time performance. Table \ref{tab:table_refine_module_results} compares the detection results of the keypoint 2D locations with and without using the refine module with the default 2D-Stage. Additionally, are also presented the results after removing the parallel connection heatmaps branch entirely.

\begin{table*}[t]
    \centering
    \caption{Importance of the refine module on the 2D-Stage accuracy and latency. Its effect is investigated by removing the refine module and obtaining the 2D keypoint locations from the keypoint heatmaps branch, and by further removing the parallel connection heatmap branch. The best results in each metric are highlighted in bold.}
    \maxsizebox{\textwidth}{!}{
    \begin{tabular}{l | c | c | c | c | c}
        Method                             & MPJPE\_2D(px) & PCK\_2D@6(\%)  & GPU latency(ms) & CPU latency(ms) & \#Params \\ \hline \hline
        Default 2D-Stage                   &\B 3.73 ± 0.04 &\B 85.27 ± 0.59 &   11.97 ± 0.03  &   37.56 ± 0.13  &   1.05M  \\
        \hspace{0.3cm}- Refine Module      &   4.37 ± 0.10 &   81.92 ± 0.71 &   11.02 ± 0.02  &   31.90 ± 0.10  &   1.01M  \\
        \hspace{0.6cm}- Connections Branch &   4.37 ± 0.10 &   81.92 ± 0.71 &\B  9.07 ± 0.04  &\B 25.89 ± 0.07  &\B 0.93M  \\
        \hline
    \end{tabular}
    }
    \label{tab:table_refine_module_results}
\end{table*}

Removing the refine module yields a 17\% increase in the MPJPE error which is traded for an also 17\% decrease in latency, which is improved further by 45\% after removing the connection branch altogether.

\subsubsection{Projection Residual}

\begin{figure}[t]
    \centering
    \begin{subfigure}[b]{0.15\textwidth}
        \centering
        \includegraphics[height=4cm]{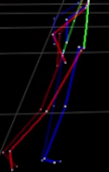}
        \caption{}
        \label{fig:results_only3d_projection_residual_raw}
    \end{subfigure}
    \begin{subfigure}[b]{0.15\textwidth}
        \centering
        \includegraphics[height=4cm]{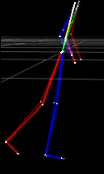}
        \caption{}
        \label{fig:results_only3d_projection_residual_corrected}
    \end{subfigure}
    \caption{\textbf{a)} 3D skeleton obtained through keypoint depth projection using the camera Pinhole model (the leg/feet keypoints were transformed to the posture camera's referential using the known extrinsic transformation). \textbf{b)} The skeleton obtained after the residual correction step of the \textit{Projection\_Residual} method.}
    \label{fig:results_only3d_raw_projection}
\end{figure}

Figure \ref{fig:results_only3d_raw_projection} shows the results obtained from directly projecting the 2D keypoint locations with depth information to 3D space using the camera Pinhole model and subsequently applying the residual correction. Table \ref{tab:table_only3d_projection_residual_results} compares these results against those obtained with the default 3D-Stage. The GT 2D locations were used to decrease the noise in the results from the 2D-Stage error.

\begin{table*}[t]
    \centering
    \caption{Default 3D-Stage lifting approach comparison with the 2D keypoint locations projection using the pixel depth and the camera Pinhole model (Projection\_Raw), and by further processing using the residual correction neural network in the Projection\_Residual method. The GT 2D keypoint locations were used as input to the regression in all methods. The best results in each metric are highlighted in bold.}
    \maxsizebox{\textwidth}{!}{
    \begin{tabular}{l | c | c | c | c | c | c}
        Method                  & MPJPE(mm)     & PA\_MPJPE(mm) & PCK@75(\%)     & CPU latency(ms) & \#Params \\ \hline \hline
        Default 3D-Stage        &\B36.65 ± 0.28 &\B28.11 ± 0.24  &\B90.64 ± 0.34 &\B 1.03 ± 0.0    &   0.29M  \\
        Projection\_Raw         &  226.0 ± 4.82 & 179.97 ± 3.11  &  21.84 ± 0.33 &   1.13 ± 0.0    &\B 0      \\
        Projection\_Residual    &  43.42 ± 0.41 &  33.56 ± 0.37  &  84.72 ± 0.45 &   2.27 ± 0.0    &   0.29M  \\
        \hline
    \end{tabular}
    }
    \label{tab:table_only3d_projection_residual_results}
\end{table*}

The raw projection method displays a larger error, above 200mm, with only 21.8\% of predictions having an error below the desired threshold of 75mm. The residual correction is capable of improving the detection substantially, to an error around 43.4mm with 84.7\% of keypoints within the detection threshold, although still higher than the lifting approaches.

\subsubsection{Only 3D-Stage}

The results from the 3D-Stage were evaluated independently from the errors of the 2D-Stage, by using the GT 2D keypoint locations as input for the regression. These results provide the maximum performance achievable given an ideal 2D-Stage and allow a less noisy comparison. Table \ref{tab:table_3d_stage_results} depicts the results and the comparison to the complete model.

\begin{table*}[t]
    \centering
    \caption{Result summary of the default 3D-Stage and lifting variants when receiving as input the GT 2D keypoint locations. The default 3D-Stage error when regressing from the 2D-Stage predictions is shown as reference. The best results in each metric are highlighted in bold.}
    \maxsizebox{\textwidth}{!}{
    \begin{tabular}{l | c | c | c | c | c}
        Method                               &   MPJPE(mm)   & PA\_MPJPE(mm) & PCK@75(\%)   & CPU latency(ms) & \#Params \\ \hline \hline
        2D-Stage                             &  &  &  &  & \\
        \hspace{0.3cm}+ Default 3D-Stage     & 44.05 ± 0.39  &  33.35 ± 0.29 & 83.03 ± 0.48 & 38.93 ± 0.13    &  1.34M   \\
        \hline
        2D-GT                                &  &  &  &  & \\
        \hspace{0.3cm}+ Default 3D-Stage     &  36.65 ± 0.28 &  28.11 ± 0.24 &  90.64 ± 0.34&\B1.03 ± 0.0     &  0.29M   \\
        \hspace{0.3cm}+ Baseline             &  36.66 ± 0.28 &  26.91 ± 0.23 &  90.75 ± 0.34&  3.98 ± 0.0     &  4.29M   \\
        \hspace{0.3cm}+ SemGCN               &\B35.22 ± 0.28 &\B26.25 ± 0.25 &\B92.59 ± 0.34&  4.38 ± 0.0     &\B0.27M   \\
        \hspace{0.3cm}+ Projection\_Residual &  43.42 ± 0.41 &  33.56 ± 0.37 &  84.72 ± 0.45&  2.27 ± 0.0     &  0.29M   \\
        \hline
    \end{tabular}
    }
    \label{tab:table_3d_stage_results}
\end{table*}

In general, all lifting methods displayed a similar level of performance (around 36mm MPJPE) with a low latency time ($<$4.38ms). The SemGCN variant obtained overall the best results, while the projection residual approach obtained once again the largest error.

An almost 8mm higher MPJPE (36.65mm) was obtained when lifting from the 2D-Stage keypoints with the default model, compared to using the 2D GT keypoints (44.05mm).

\subsubsection{Temporal Model}

The results obtained using the spatio-temporal stages (Section \ref{sec:model_variant_spatiotemporal}) were compared next, starting with the 2D-Stage in Table \ref{tab:results_temporal_2dstage} and the 3D-Stage (from 2D keypoint predictions and GT in Table \ref{tab:results_temporal_3dstage}.

\begin{table*}[t]
    \centering
    \caption{2D-Stage results obtained using the temporal model with 4 sequential frames, compared to the default single-frame version.}
    \maxsizebox{\textwidth}{!}{
    \begin{tabular}{l | c | c | c | c | c}
        Method               & MPJPE\_2D(px) & PCK\_2D@6(\%)  & GPU latency(ms) & CPU latency(ms) & \#Params \\ \hline \hline
        Single(1 frame)      &\B 3.73 ± 0.04 &\B 85.27 ± 0.59 &\B 11.97 ± 0.03  &\B 37.56 ± 0.13  &\B 1.05M  \\
        Sequential(4 frames) &   4.13 ± 0.06 &   83.41 ± 0.63 &   51.13 ± 0.05  &  554.08 ± 2.07  &   1.40M  \\
        \hline
    \end{tabular}
    }
    \label{tab:results_temporal_2dstage}
\end{table*}

The 2D spatio-temporal model obtained slightly worse detection results, with a 2D MPJPE of 4.13 pixels, while being noticeably slower to compute, especially on the CPU, taking 554ms to process 4 frames, compared to the single-frame counterpart.

\begin{table*}[t]
    \centering
    \caption{3D results obtained using the temporal model with 4 sequential frames, compared to the proposed single-frame version, from the 2D keypoint location predictions of the 2D-Stage and GT. The best results in each metric are highlighted in bold.}
    \maxsizebox{\textwidth}{!}{
    \begin{tabular}{l | c | c | c | c | c | c}
        Method                               &   MPJPE(mm)   & PA\_MPJPE(mm) & PCK@75(\%)    & GPU latency(ms) & CPU latency(ms) & \#Params \\ \hline \hline
        2D-Stage                             &  &  &  &  &  & \\
        \hspace{0.3cm}+ Single(1 frame)      &\B44.05 ± 0.39 &\B33.35 ± 0.29 &\B83.03 ± 0.48 &\B13.43 ± 0.03   &\B 38.93 ± 0.13  &\B 1.34M   \\
        \hspace{0.3cm}+ Sequential(4 frames) &  46.72 ± 0.56 &  34.93 ± 0.41 &  82.61 ± 0.51 &  52.50 ± 0.06   &   554.86 ± 2.45 &   1.98M   \\
        \hline
        2D-GT                            &  &  &  &  &  & \\
        \hspace{0.3cm}+ Single(1 frame)      &  36.65 ± 0.28 &\B28.11 ± 0.24 &  90.64 ± 0.34 &\B1.31 ± 0.00    &\B 1.03 ± 0.0    &\B 0.29M   \\
        \hspace{0.3cm}+ Sequential(4 frames) &\B36.49 ± 0.29 &  28.79 ± 0.23 &\B91.01 ± 0.36 &  1.77 ± 0.00    &   1.60 ± 0.0    &   0.58M   \\
        \hline
    \end{tabular}
    }
    \label{tab:results_temporal_3dstage}
\end{table*}

The 3D-Stage yielded similar performance to the default single-frame version, being marginally worse when using the predictions from the 2D-Stage, given the slightly worse results obtained by the sequential 2D-Stage. The latency of the temporal 3D-Stage is slightly higher than the single-frame model but it is overall faster since 4 frames are processed simultaneously.

Additionally, the sequential predictions yield temporally coherent results with no jittering when applied to video streams, without requiring the post-processing temporal filter introduced in the single-frame counterpart, while introducing no response delay.

\section{Discussion}
\label{sec:discussion}

We presented a working real-time full body pose estimation solution for the ASBGo smart, able to extract a compact body configuration representation, from the RGB+D cameras stream, which can be used in downstream tasks in activity and intent recognition, enable human-robot interaction applications and facilitate extraction of multiple gait and posture metrics for clinical evaluation, with further processing. Multiple benchmarks and ablations studies were performed to evaluate and better understand the model performance, as well as exploring competing approaches.

The GT labels obtained through the acquisition method are not perfect since the Xsens data contain no visual correlation with the camera streams during acquisition, resulting in positional offsets, which is worsened by compounding errors when relating both the referentials (Xsens calibration errors, extrinsic camera calibration, users wrist position on the handles). These errors can result in bad samples which do not align visually with the limbs on some frames.

Depth data could be used to pre-process the frames to segment the user by applying geometric and threshold operations, before feeding them to the models as in \cite{Paulo2017AIWALKER}. However, this approach was not followed since the neural networks are capable of parsing this information from the raw frames, the additional processing overhead, and the introduction of failure cases which would decrease the models' robustness.

The location objective is rather ambiguous below a certain threshold, since many locations in the neighborhood could be considered correct for each keypoint. Moreover, due to the presence of some noise in the GT labels, it would be impossible to obtain no error. The wrist keypoints displayed the lowest errors, given the low variability as the subjects are required to be grabbing the walker handles at all times, while, the feet keypoints displayed the largest errors and presence of outliers, given the amount of movement and possibility of occlusions.

The overall results indicate a similar performance across model variants both for the 2D and 3D-Stages ($\approx$ 3.73px and 44.0mm). This might indicate performance saturation for the task, given the amount of data and the noise present in the dataset.

Despite its simplicity, the low-pass filter removed high-frequency jitter (Figure \ref{fig:results_3d_saggital_filter}), coming from the raw single-frame model predictions. This improves temporal coherence and leads to more accurate joint kinematics, required for downstream tasks, while adding an insignificant amount of latency overhead. Nevertheless, this comes at the cost of attenuation of natural dynamics and introduction of response delay.

The model is quite robust to corruption in either the inputs (Figure \ref{fig:results_corrupted_inputs}), being able to work with no depth or no image information, having to rely entirely on the remaining input feature. In these cases, the prediction confidence is lower, with also lower temporal consistency across frames. Although an improbable situation during rehabilitation settings, it shows some of the potentials of using learning-based methods from a robustness standpoint.

Both 2D keypoint and connection heatmaps features, obtained from the intermediate objective, displayed the expected and interpretable Gaussian shape (Figure \ref{fig:results_only2d_detection_good}). This indicates that the model has correctly learned the features from the secondary regularization loss, since these are complementary to the keypoint detection objective.

The EfficientNet-lite0 feature extraction backbone displayed by far the lowest computation time in the CPU (Table \ref{tab:table_backbones_results}), being faster than the ResNet18 baseline by 2.5 times and the ResNet50 by 6.4 times. Interestingly, the ResNet backbones offer much closer run-time performance when tested on the GPU. This might be explained by the fact that bigger, but simpler convolution operations are used, optimized to run on the GPU which processes large operations simultaneously. On the other hand, the EfficientNet backbone separates each convolution block into multiple small steps which are run sequentially introducing some overhead. When it comes to inference on the CPU, the great reduction in operations in the EfficientNet modules is greater than the overhead for the multiple sequential calls, yielding noticeably faster run-times, making it more adequate for the walker hardware. For applications targeting the GPU, the ResNet or its variants could be a better option.

The shallow refine module is capable of aggregating information to produce better results overall (Table \ref{tab:table_refine_module_results}), by also using connection cues from the connection heatmaps branch. These performance improvements come nonetheless at the cost of an increase around 45\% in latency, which was considered acceptable since it was still within the performance requirements imposed.

The baseline lifting approach without depth information achieved similar results as the one with depth, both in relative and absolute spaces (Table \ref{tab:table_complete_3D_results}). This means that depth is not required for HPE in this task, possibly since the poses have low variance, ambiguous poses are uncommon and the wrist positions are almost constant, while the depth information might not be too reliable since only a single noisy point is considered for each keypoint. This indicates that future frameworks based solely on RGB camera data would be possible, further decreasing the cost of hardware.

The 3D keypoints predicted by the model, despite showing similar trajectories (Figure \ref{fig:results_3d_saggital_filter}) to those generated by the Xsens GT, still displayed some constant positional offset and failure to capture the full range of motion on others. It might also predict incorrect limb lengths, that do not correspond to the subject's anthropometric data.

Although the leg and feet keypoints are detected on a different image reference frame from the torso, the lifting models are capable of internally finding a way to relate the information without the need of explicitly providing the extrinsic transformation between camera frames.

Furthermore, it was shown (Table \ref{tab:table_only3d_projection_residual_results}) the lifting approach yields better results than the explicit projection method with residual correction. This could be explained by the fact that the lifting method always considers the 2D location information along with the noisy depth values to produce the transformation. On the other hand, the projection method, although having a simplified task since the referential transformations are computed externally, in practice, is faced with frequent cases where one or more keypoints contain incorrect or no information due to bad projection (dead pixels, background pixel selection, different body thickness across subjects and body parts).

Unexpectedly, the complete temporal model performed worse when benchmarked against the single-frame counterparts (Table \ref{tab:results_temporal_3dstage}). Multiple reasons might have contributed to this. Namely, the use of a backbone pre-trained with longer temporal sequences, the harder optimization process given the more complex task of also relating temporal information, the use of a model with higher capacity given the same amount of data which might have resulted in some over-fitting. Nevertheless, these create temporally coherent results, without the need for any post-processing, preferable when dealing with sequential data. Unfortunately, the 3D convolutions added a significant amount of latency ($>$500ms) especially in the CPU, making them impractical for the walker.

The model worked as expected on the smart walker (Figure \ref{fig:results_deployed_3d}), running in real-time while displaying similar detection performance (qualitatively speaking) to the dataset test samples, when dealing with common walking situations used for training and complying with the requirements imposed in Section \ref{sec:requirements}. However, the performance degraded when presented with situations outside the training distribution, implying that more diverse data is needed to train a model capable of fully performing in real-world situations. Moreover, increased latency and response delay were reported, due to the asynchronous nature of the underlying ROS system in the equipment.

Some model ideas had to be changed or dropped entirely, as these were not correctly supported by the ONNX framework (and also most alternatives available), and thus not fit for the deployment scheme used. This included the use of: \textbf{i)} SE attention blocks proposed by \cite{Hu2017senets} for the 2D-Stage, tried initially with promising results in terms of performance with minimal effect on latency; \textbf{ii)} EfficientNet (non-lite) versions, which also used the SE modules; \textbf{iii)} SemGCN \cite{Zhao2019semgcn} modules for the 2D to 3D lifting.

The proposed framework, based on learning algorithms, allows extracting a compact representation of the full human body, directly from inexpensive cameras and adaptable (given some training data) to multiple setup configurations. This is in opposition to competing smart walker solutions that rely on custom dedicated hardware, must be carefully tuned and are only capable of monitoring a small set of metrics. Nevertheless, given the "black-box" nature of DL algorithms, weaker guarantees about the validity of the predictions are possible, since these do not directly represent physical measurements in the inputs (e.g., depth), as in classical methods. The models are trained to produce confident predictions for data that fall inside the distribution used for training and will produce unreasonable predictions for data that falls outside. This is still an open issue in the DL field and thus, we emphasise the need to fully benchmark deployed solutions prior to real-world usage.

Extensive hyper-parameter optimization was not performed, and better results could have been achieved by hyper-parameter search. However, given that a general framework is proposed, and the computational resources available, we considered more beneficial to perform a fair comparison between different approaches, using reasonable configurations based on values commonly found in the DL literature.

The use of the walker's relatively weak computing power, coupled with the real-time requirements, limited the use of more accurate, pretrained NN implementations (\textit{e.g., OpenPose \cite{Cao2017openpose}} - Table \ref{tab:table_backbones_results}). The use of such models would be possible if hardware upgrades to the walker (\textit{e.g.,} by adding a GPU) were possible. Offloading computations to an external station connected through WiFi was also considered, but discarded, given to the introduction of communication delays and failure points due to bad connection.

Assuming that the subject's hands were always placed on the walker handles limited the pool of movements available for analysis. However, this is consistent with use by patients in gait rehabilitation, for balance and stability purposes. Alternatively, visual markers could be used on both wrists to determine the position offset on all frames. However, the added complexity (i.e. extra setup/processing steps, additional hardware) was deemed unreasonable for this application.

The dataset also contains relatively low variability in terms of poses and their locations on the image frames, limiting its application to general HPE problems. Data augmentation techniques were applied to the 2D-Stage to decrease visually and positional overfitting to the common keypoint locations on the image frames, while also collecting additional 13k frames with irregular walking data which brought some improvements. Nevertheless, these were not sufficient to combat model overfitting to the overwhelming amount of common keypoint locations during walking. This made it unreasonable to explore more complex models, and fully train the 2D-Stage, as it would easily overfit. A possible solution to mitigate this problem would involve pretraining each stage or the full model on a general HPE dataset \cite{Joo2017cmupanoptic} with similar data modalities (image, depth) and then fine-tune the last few layers on this dataset.

\section{Conclusion}
\label{sec:conclusion}

A novel full-body pose estimation solution for the ASBGo smart walker was developed. It is able to extract a compact body representation from two camera streams, which can be used for downstream tasks in patient monitoring and enable human-in-the-loop control strategies.


A strong focus was given to developing methods which can run in real-time. A two-staged approach was favored, running at 25fps (40ms) on the constrained CPU of the walker, in rehabilitation settings. The first stage focuses on the harder problem of 2D detection, using a fast FCN detector to locate the subject's keypoints on the image. Next, a residual regression module at the second stage, infers the depth of each keypoint relative to the camera. A MPJPE of 44mm was reported for the complete model. Temporal consistency was achieved through post-processing with a low-pass adaptive filter.

Multiple ablation studies were conducted, to investigate components relevant to the final model's performance. An accurate and fast backbone was considered necessary to obtain good results at low latency, while the 3D regression network was less restricted by runtime latency. 
Alternative ideas were benchmarked against the proposed method, following literature suggestions, but no significant improvements over the default method were found, possibly due to performance saturation on the dataset. Nevertheless, the results clearly show the advantages of using a learned projection approach compared to standard depth projection, as used in current smart walker solutions. Moreover, the temporal NN model was capable of producing temporally consistent results without post-processing, although not in real-time.

Promising results were obtained on healthy participants. Nonetheless, more data, with also higher variability, should be collected with gait impaired subjects to access the framework's true performance as a rehabilitation tool in real-world scenarios. An extension to the dataset is thus planned, using the same acquisition setup, during real rehabilitation sessions. It will be used to further training the models and allow validation in clinical scenarios.

Future Improvements to the framework include: \textit{i)} the combination of a kinematic model with the predictions from the NN, to introduce prior knowledge and improving robustness, by only allowing the production of anthropomorphically valid outputs; \textit{ii)} the combination of a single-frame 2D-Stage with a temporal 3D-Stage, to allow temporally consistent 3D outputs with low computational time and no response delay.





\section*{Acknowledgements}
This work has been supported by the FEDER funds through the COMPETE 2020 - Programa Operacional Competitividade e Internacionalização (POCI) and P2020 with the Reference Project EML under Grant POCI-01-0247-FEDER-033067, and by FCT national funds, under the national support to R\&D units grant, through the reference project UIDB/04436/2020 and UIDP/04436/2020.

\section*{Declaration of Competing Interests}
The authors declare that they have no known competing financial interests or personal relationships that could have appeared to influence the work reported in this paper.

\ifCLASSOPTIONcaptionsoff
  \newpage
\fi

\bibliographystyle{IEEEtran}  
\bibliography{References.bib}   

\begin{thebibliography}{10}
\providecommand{\url}[1]{#1}
\csname url@samestyle\endcsname
\providecommand{\newblock}{\relax}
\providecommand{\bibinfo}[2]{#2}
\providecommand{\BIBentrySTDinterwordspacing}{\spaceskip=0pt\relax}
\providecommand{\BIBentryALTinterwordstretchfactor}{4}
\providecommand{\BIBentryALTinterwordspacing}{\spaceskip=\fontdimen2\font plus
\BIBentryALTinterwordstretchfactor\fontdimen3\font minus
  \fontdimen4\font\relax}
\providecommand{\BIBforeignlanguage}[2]{{%
\expandafter\ifx\csname l@#1\endcsname\relax
\typeout{** WARNING: IEEEtran.bst: No hyphenation pattern has been}%
\typeout{** loaded for the language `#1'. Using the pattern for}%
\typeout{** the default language instead.}%
\else
\language=\csname l@#1\endcsname
\fi
#2}}
\providecommand{\BIBdecl}{\relax}
\BIBdecl

\bibitem{WHO2011}
WHO, \emph{World Report on Disability}.\hskip 1em plus 0.5em minus 0.4em\relax
  World Health Organization, 2011.

\bibitem{Mikolajczyk2018}
T.~Mikolajczyk, I.~Ciobanu, D.~I. Badea, A.~Iliescu, S.~Pizzamiglio,
  T.~Schauer, T.~Seel, P.~L. Seiciu, D.~L. Turner, and M.~Berteanu, ``{Advanced
  technology for gait rehabilitation: An overview},'' \emph{Advances in
  Mechanical Engineering}, vol.~10, no.~7, pp. 1--19, 2018.

\bibitem{Jonsdottir2017}
J.~Jonsdottir and M.~Ferrarin, \emph{Gait Disorders in Persons After
  Stroke}.\hskip 1em plus 0.5em minus 0.4em\relax Cham: Springer International
  Publishing, 2017, pp. 1--11.

\bibitem{Johnson2016}
W.~Johnson, O.~Onuma, M.~Owolabi, and S.~Sachdev, ``Stroke: A global response
  is needed,'' \emph{Bulletin of the World Health Organization}, vol.~94,
  no.~9, pp. 634A--635A, 2016.

\bibitem{Moreira2019}
R.~Moreira, J.~Alves, A.~Matias, and C.~P. Santos, \emph{{Smart and Assistive
  Walker – ASBGo: Rehabilitation Robotics: A Smart– Walker to Assist Ataxic
  Patients}}.\hskip 1em plus 0.5em minus 0.4em\relax Springer Nature
  Switzerland AG, 2019, pp. 37--68.

\bibitem{Martins2012smartwalkers}
M.~M. Martins, C.~P. Santos, A.~Frizera-Neto, and R.~Ceres, ``Assistive
  mobility devices focusing on smart walkers: Classification and review,''
  \emph{Robotics and Autonomous Systems}, vol.~60, no.~4, pp. 548 -- 562, 2012.

\bibitem{Kidzinski2020analysis}
{\L}.~Kidzi{\'n}ski, B.~Yang, J.~Hicks, A.~Rajagopal, S.~Delp, and M.~Schwartz,
  ``Deep neural networks enable quantitative movement analysis using
  single-camera videos,'' \emph{Nature Communications}, vol.~11, no.~1, 2020.

\bibitem{Chen2020Pose}
Y.~Chen, Y.~Tian, and M.~He, ``Monocular human pose estimation: A survey of
  deep learning-based methods,'' \emph{Computer Vision and Image
  Understanding}, vol. 192, p. 102897, 2020.

\bibitem{neto2015walkers}
A.~F. Neto, A.~Elias, C.~Cifuentes, C.~Rodriguez, T.~Bastos, and R.~Carelli,
  ``Smart walkers: Advanced robotic human walking-aid systems,'' in
  \emph{Intelligent Assistive Robots}.\hskip 1em plus 0.5em minus 0.4em\relax
  Springer, 2015, pp. 103--131.

\bibitem{Paulo2017AIWALKER}
J.~{Paulo}, P.~{Peixoto}, and U.~J. {Nunes}, ``Isr-aiwalker: Robotic walker for
  intuitive and safe mobility assistance and gait analysis,'' \emph{IEEE
  Transactions on Human-Machine Systems}, vol.~47, no.~6, pp. 1110--1122, 2017.

\bibitem{Angeloni1994gaitfreq}
C.~{Angeloni}, P.~O. {Riley}, and D.~E. {Krebs}, ``Frequency content of whole
  body gait kinematic data,'' \emph{IEEE Transactions on Rehabilitation
  Engineering}, vol.~2, no.~1, pp. 40--46, 1994.

\bibitem{Frizera2011Simbiosis}
A.~Frizera-Neto, R.~Ceres, E.~Rocon, and J.~L. Pons, ``Empowering and assisting
  natural human mobility: The simbiosis walker,'' \emph{International Journal
  of Advanced Robotic Systems}, vol.~8, no.~3, p.~29, 2011.

\bibitem{Sierra2019AGoRA}
S.~D. Sierra~M, M.~Garz{\'o}n, M.~Munera, C.~A. Cifuentes \emph{et~al.},
  ``Human-robot-environment interaction interface for smart walker assisted
  gait: Agora walker,'' \emph{Sensors}, vol.~19, no.~13, p. 2897, 2019.

\bibitem{mou2012CAIROW}
W.-H. Mou, M.-F. Chang, C.-K. Liao, Y.-H. Hsu, S.-H. Tseng, and L.-C. Fu,
  ``Context-aware assisted interactive robotic walker for parkinson's disease
  patients,'' in \emph{2012 IEEE/RSJ International Conference on Intelligent
  Robots and Systems}.\hskip 1em plus 0.5em minus 0.4em\relax IEEE, 2012, pp.
  329--334.

\bibitem{mehta2017vnect}
D.~Mehta, S.~Sridhar, O.~Sotnychenko, H.~Rhodin, M.~Shafiei, H.-P. Seidel,
  W.~Xu, D.~Casas, and C.~Theobalt, ``Vnect: Real-time 3d human pose estimation
  with a single rgb camera,'' \emph{ACM Transactions on Graphics (TOG)},
  vol.~36, no.~4, pp. 1--14, 2017.

\bibitem{Groos2020efficient}
D.~Groos, H.~Ramampiaro, and E.~A. Ihlen, ``Efficientpose: Scalable
  single-person pose estimation,'' \emph{Applied Intelligence}, 2020.

\bibitem{Cao2017openpose}
Z.~Cao, T.~Simon, S.-E. Wei, and Y.~Sheikh, ``Realtime multi-person 2d pose
  estimation using part affinity fields,'' in \emph{Proceedings of the IEEE
  conference on computer vision and pattern recognition}, 2017, pp. 7291--7299.

\bibitem{Moccia2020preterm}
S.~Moccia, L.~Migliorelli, V.~Carnielli, and E.~Frontoni, ``Preterm infants’
  pose estimation with spatio-temporal features,'' \emph{IEEE Transactions on
  Biomedical Engineering}, vol.~67, no.~8, p. 2370–2380, 2020.

\bibitem{Achilles2016blanket}
F.~Achilles, A.-E. Ichim, H.~Coskun, F.~Tombari, S.~Noachtar, and N.~Navab,
  ``Patient mocap: Human pose estimation under blanket occlusion for hospital
  monitoring applications,'' in \emph{Medical Image Computing and
  Computer-Assisted Intervention (MICCAI 2016)}, S.~Ourselin, L.~Joskowicz,
  M.~R. Sabuncu, G.~Unal, and W.~Wells, Eds.\hskip 1em plus 0.5em minus
  0.4em\relax Cham: Springer International Publishing, 2016, pp. 491--499.

\bibitem{Quigley2009Ros}
M.~Quigley, B.~Gerkey, K.~Conley, J.~Faust, T.~Foote, J.~Leibs, E.~Berger,
  R.~Wheeler, and A.~Ng, ``Ros: an open-source robot operating system,'' in
  \emph{Proc. of the IEEE Intl. Conf. on Robotics and Automation (ICRA)
  Workshop on Open Source Robotics}, Kobe, Japan, May 2009.

\bibitem{Newell2016hourglass}
A.~Newell, K.~Yang, and J.~Deng, ``Stacked hourglass networks for human pose
  estimation,'' in \emph{Computer Vision -- ECCV 2016}, B.~Leibe, J.~Matas,
  N.~Sebe, and M.~Welling, Eds.\hskip 1em plus 0.5em minus 0.4em\relax Cham:
  Springer International Publishing, 2016, pp. 483--499.

\bibitem{tompson2015heatmaps}
J.~Tompson, R.~Goroshin, A.~Jain, Y.~LeCun, and C.~Bregler, ``Efficient object
  localization using convolutional networks,'' in \emph{Proceedings of the IEEE
  conference on computer vision and pattern recognition}, 2015, pp. 648--656.

\bibitem{Nibali2019margipose}
A.~Nibali, Z.~He, S.~Morgan, and L.~Prendergast, ``3d human pose estimation
  with 2d marginal heatmaps,'' in \emph{2019 IEEE Winter Conference on
  Applications of Computer Vision (WACV)}.\hskip 1em plus 0.5em minus
  0.4em\relax IEEE, 2019, pp. 1477--1485.

\bibitem{Martinez2017baseline}
J.~Martinez, R.~Hossain, J.~Romero, and J.~J. Little, ``A simple yet effective
  baseline for 3d human pose estimation,'' in \emph{Proceedings of the IEEE
  International Conference on Computer Vision}, 2017, pp. 2640--2649.

\bibitem{Wei2016cpm}
S.-E. Wei, V.~Ramakrishna, T.~Kanade, and Y.~Sheikh, ``Convolutional pose
  machines,'' in \emph{Proceedings of the IEEE conference on Computer Vision
  and Pattern Recognition}, 2016, pp. 4724--4732.

\bibitem{Sun2018integral}
X.~Sun, B.~Xiao, F.~Wei, S.~Liang, and Y.~Wei, ``Integral human pose
  regression,'' in \emph{Proceedings of the European Conference on Computer
  Vision (ECCV)}, 2018, pp. 529--545.

\bibitem{artacho2020unipose}
B.~Artacho and A.~Savakis, ``Unipose: Unified human pose estimation in single
  images and videos,'' in \emph{Proceedings of the IEEE/CVF Conference on
  Computer Vision and Pattern Recognition}, 2020, pp. 7035--7044.

\bibitem{Tan2019efficientnet}
M.~Tan and Q.~V. Le, ``Efficientnet: Rethinking model scaling for convolutional
  neural networks,'' in \emph{Proceedings of the 36th International Conference
  on Machine Learning, {ICML} 2019, 9-15 June 2019, Long Beach, California,
  {USA}}, ser. Proceedings of Machine Learning Research, K.~Chaudhuri and
  R.~Salakhutdinov, Eds., vol.~97.\hskip 1em plus 0.5em minus 0.4em\relax
  {PMLR}, 2019, pp. 6105--6114.

\bibitem{Chen2018deeplab}
L.-C. Chen, Y.~Zhu, G.~Papandreou, F.~Schroff, and H.~Adam, ``Encoder-decoder
  with atrous separable convolution for semantic image segmentation,'' in
  \emph{Proceedings of the European conference on computer vision (ECCV)},
  2018, pp. 801--818.

\bibitem{Lin2017focal}
T.-Y. Lin, P.~Goyal, R.~Girshick, K.~He, and P.~Doll{\'a}r, ``Focal loss for
  dense object detection,'' in \emph{Proceedings of the IEEE international
  conference on computer vision}, 2017, pp. 2980--2988.

\bibitem{Gonzalez2020residual}
A.~Mart{\'{\i}}nez-Gonz{\'{a}}lez, M.~Villamizar, O.~Can{\'{e}}vet, and J.-M.
  Odobez, ``Residual pose: A decoupled approach for depth-based 3d human pose
  estimation,'' in \emph{IEEE/RSJ International Conference on Intelligent
  Robots and Systems}, 2020.

\bibitem{casiez2012oneeurofilter}
G.~Casiez, N.~Roussel, and D.~Vogel, ``1€ filter: a simple speed-based
  low-pass filter for noisy input in interactive systems,'' in
  \emph{Proceedings of the SIGCHI Conference on Human Factors in Computing
  Systems}, 2012, pp. 2527--2530.

\bibitem{Beaman2010speed}
C.~Beaman, C.~Peterson, R.~Neptune, and S.~Kautz, ``Differences in
  self-selected and fastest-comfortable walking in post-stroke hemiparetic
  persons,'' \emph{Gait \& Posture}, vol.~31, no.~3, pp. 311 -- 316, 2010.

\bibitem{Shorten2019augm}
C.~Shorten and T.~Khoshgoftaar, ``A survey on image data augmentation for deep
  learning,'' \emph{Journal of Big Data}, vol.~6, pp. 1--48, 2019.

\bibitem{Sarandi2018augm}
I.~S{\'a}r{\'a}ndi, T.~Linder, K.~O. Arras, and B.~Leibe, ``How robust is 3d
  human pose estimation to occlusion?'' in \emph{IROS Workshop - Robotic
  Co-workers 4.0}, 2018.

\bibitem{Zhao2019semgcn}
L.~Zhao, X.~Peng, Y.~Tian, M.~Kapadia, and D.~N. Metaxas, ``Semantic graph
  convolutional networks for 3d human pose regression,'' in \emph{Proceedings
  of the IEEE Conference on Computer Vision and Pattern Recognition}, 2019, pp.
  3425--3435.

\bibitem{Pavllo2019temporal}
D.~Pavllo, C.~Feichtenhofer, D.~Grangier, and M.~Auli, ``3d human pose
  estimation in video with temporal convolutions and semi-supervised
  training,'' in \emph{Conference on Computer Vision and Pattern Recognition
  (CVPR)}, 2019, pp. 7745--7754.

\bibitem{Sandler2018mobilenet2}
M.~{Sandler}, A.~{Howard}, M.~{Zhu}, A.~{Zhmoginov}, and L.~{Chen},
  ``Mobilenetv2: Inverted residuals and linear bottlenecks,'' in \emph{2018
  IEEE/CVF Conference on Computer Vision and Pattern Recognition}, 2018, pp.
  4510--4520.

\bibitem{Hu2017senets}
J.~Hu, L.~Shen, and G.~Sun, ``Squeeze-and-excitation networks,'' in
  \emph{Proceedings of the IEEE conference on computer vision and pattern
  recognition}, 2018, pp. 7132--7141.

\bibitem{Joo2017cmupanoptic}
H.~Joo, T.~Simon, X.~Li, H.~Liu, L.~Tan, L.~Gui, S.~Banerjee, T.~S. Godisart,
  B.~Nabbe, I.~Matthews, T.~Kanade, S.~Nobuhara, and Y.~Sheikh, ``Panoptic
  studio: A massively multiview system for social interaction capture,''
  \emph{IEEE Transactions on Pattern Analysis and Machine Intelligence}, 2017.

\end{thebibliography}

\end{document}